\documentclass[pdflatex,sn-mathphys,iicol]{sn-jnl}

\jyear{2023}%

\theoremstyle{thmstyleone}%
\theoremstyle{thmstyletwo}%

\theoremstyle{thmstylethree}%

\usepackage{soul}
\soulregister\cite7
\soulregister\ref7
\usepackage{color}
\usepackage{hyperref} 
\hypersetup{
	colorlinks=true,
	urlcolor=magenta,
}

\begin{document}
\title[Article Title]{CBNet: A Plug-and-Play Network for Segmentation-Based Scene Text Detection}


\author*[1]{\fnm{Xi} \sur{Zhao}}\email{xizhao24@outlook.com}
\author[1]{\fnm{Wei} \sur{Feng}}
\author[1]{\fnm{Zheng} \sur{Zhang}}
\author[1]{\fnm{Jingjing} \sur{Lv}}
\author[1]{\fnm{Xin} \sur{Zhu}}
\author[1]{\fnm{Zhangang} \sur{Lin}}
\author[1]{\fnm{Jinghe} \sur{Hu}}
\author[1]{\fnm{Jingping} \sur{Shao}}

\affil*[1]{\orgdiv{Marketing \& Commercialization Center}, \orgname{JD.com}, \orgaddress{\city{Beijing}, \postcode{101111}, \country{China}}}


\abstract{Recently, segmentation-based methods are quite popular in scene text detection, which mainly contain two steps: text kernel segmentation and expansion. However, the segmentation process only considers each pixel independently, and the expansion process is difficult to achieve a favorable accuracy-speed trade-off. In this paper, we propose a \textbf{C}ontext-aware and \textbf{B}oundary-guided \textbf{N}etwork (CBN) to tackle these problems. In CBN, a basic text detector is first used to predict initial segmentation results. Then, we propose a context-aware module to enhance text kernel feature representations, which considers both global and local contexts. Finally, we introduce a boundary-guided module to expand enhanced text kernels adaptively with only the pixels on the contours, which not only obtains accurate text boundaries but also keeps high speed, especially on high-resolution output maps. In particular, with a lightweight backbone, the basic detector equipped with our proposed CBN achieves state-of-the-art results on several popular benchmarks, and our proposed CBN can be plugged into several segmentation-based methods. Code is available at \url{https://github.com/XiiZhao/cbn.pytorch}.}




\maketitle

\section{Introduction}
Scene text detection has received increasing attention in both academia and industry due to its wide real-world applications, such as text retrieval, scene parsing, video understanding, and automatic driving. Benefited from the powerful representation of deep learning, scene text detection has achieved great progress in the past few years~\cite{wang2019shape,liao2020real,wang2019efficient,dai2021progressive,zhang2021adaptive,liao2022real,tang2022few}. However, scene text detection is still a challenging task because of the variety of text shapes, orientations, and scales.

\begin{figure*}[t!]
	\centering
	\begin{subfigure}[b]{.48\textwidth}
		\centering
		\includegraphics[width=\textwidth]{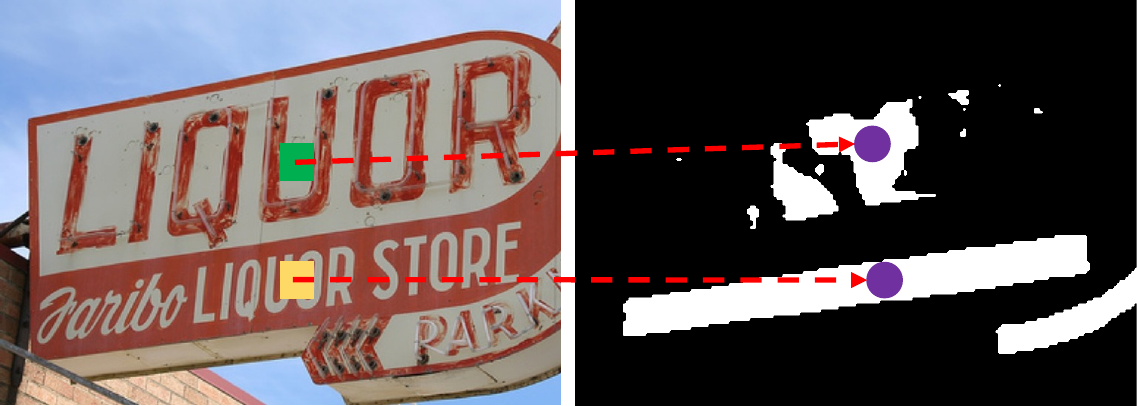}
		\caption{}
		\label{subfig:baseline}
	\end{subfigure}
	\hspace{2.5pt}	
	\begin{subfigure}[b]{.48\textwidth}
		\centering
		\includegraphics[width=\textwidth]{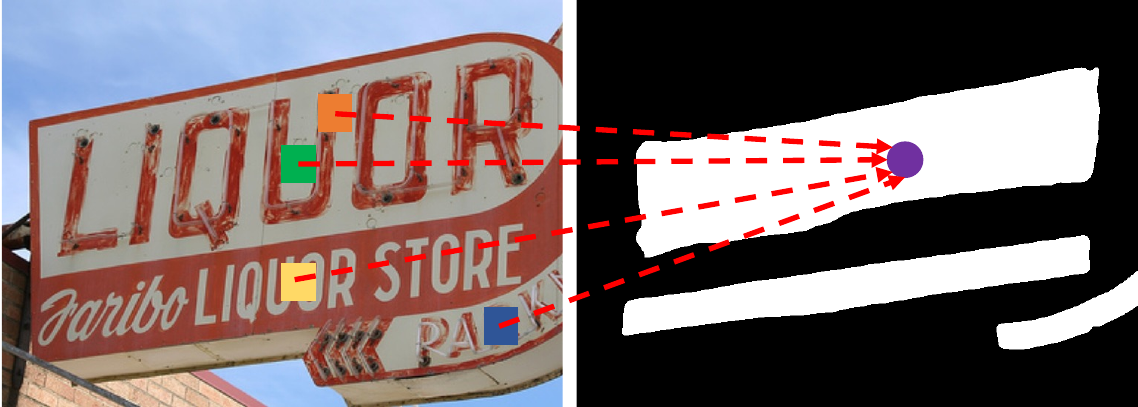}
		\caption{}
		\label{subfig:cbn}
	\end{subfigure}
	\begin{subfigure}[b]{.48\textwidth}
		\centering
		\includegraphics[width=\textwidth]{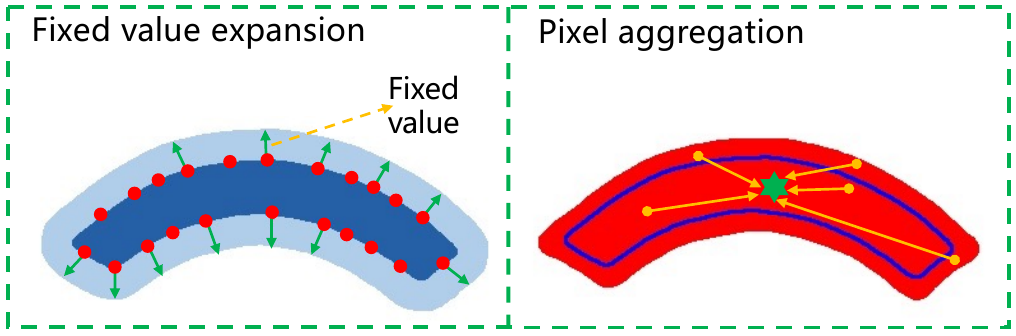}
		\caption{}
		\label{subfig:basepost}
	\end{subfigure}
    \hspace{2.5pt}
	\begin{subfigure}[b]{.48\textwidth}
	\centering
	\includegraphics[width=\textwidth]{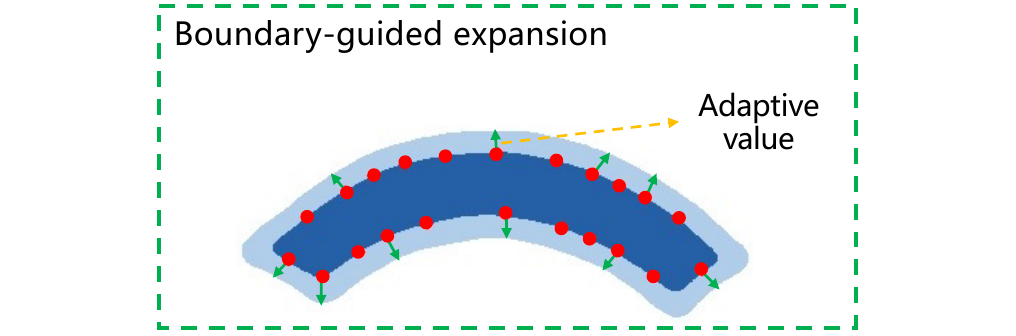}
	\caption{}
	\label{subfig:bgpost}
    \end{subfigure}
	\caption{Most existing segmentation-based methods contain two steps: text kernel segmentation and expansion. (a) and (b) show that considering the relationship between pixels can improve segmentation results. The color masks in (a) and (b) represent the context of pixels. (c) and (d) show that using the points on the contour to expand the text kernel adaptively achieves a favorable accuracy-speed trade-off. The red pixels in (c) and (d) represent the pixels that need to participate in the kernel expansion.}
	\label{fig:intro}
\end{figure*}

To address this challenge, some regression-based methods~\cite{liao2017textboxes,liu2017deep,zhou2017east,ma2018arbitrary,liao2018textboxes++} employ rectangles or quadrangles to locate text instances. However, these methods fail to handle arbitrary shape text instances. To flexibly represent the shape of text, segmentation-based methods~\cite{zhang2016multi,yao2016scene,lyu2018mask,deng2018pixellink,long2018textsnake,wang2019shape,xie2019scene,xu2019textfield,zhu2021textmountain} usually obtain text boundaries through text kernel segmentation and expansion. As shown in Fig.~\ref{fig:intro}, the text kernel segmentation step firstly shrinks original text regions, which can help to distinguish adjacent text instances. After that, the text kernel expansion step restores the shrunk text region to the complete contour, such as fixed value expansion~\cite{lyu2018mask,xie2019scene,liao2020real} or pixel aggregation~\cite{deng2018pixellink,long2018textsnake,xu2019textfield,wang2019efficient,wang2019shape,zhu2021textmountain}. Although segmentation-based methods are more robust to the diversity of text shapes, there are still some drawbacks in these two steps. On the one hand, each pixel in the text kernel is predicted independently as shown in Fig.~\ref{subfig:baseline}, which ignores the relationship between pixels. On the other hand, as shown in Fig.~\ref{subfig:basepost}, using fixed values to expand is easily affected by the quality of the text kernel. Meanwhile, pixel aggregation needs to operate on all positive pixels, which is unfriendly to efficiency when outputting high-resolution maps. Therefore, segmentation-based text detectors still need to be carefully designed.

\begin{figure*}[t]
	\centering
	\includegraphics[width=\linewidth]{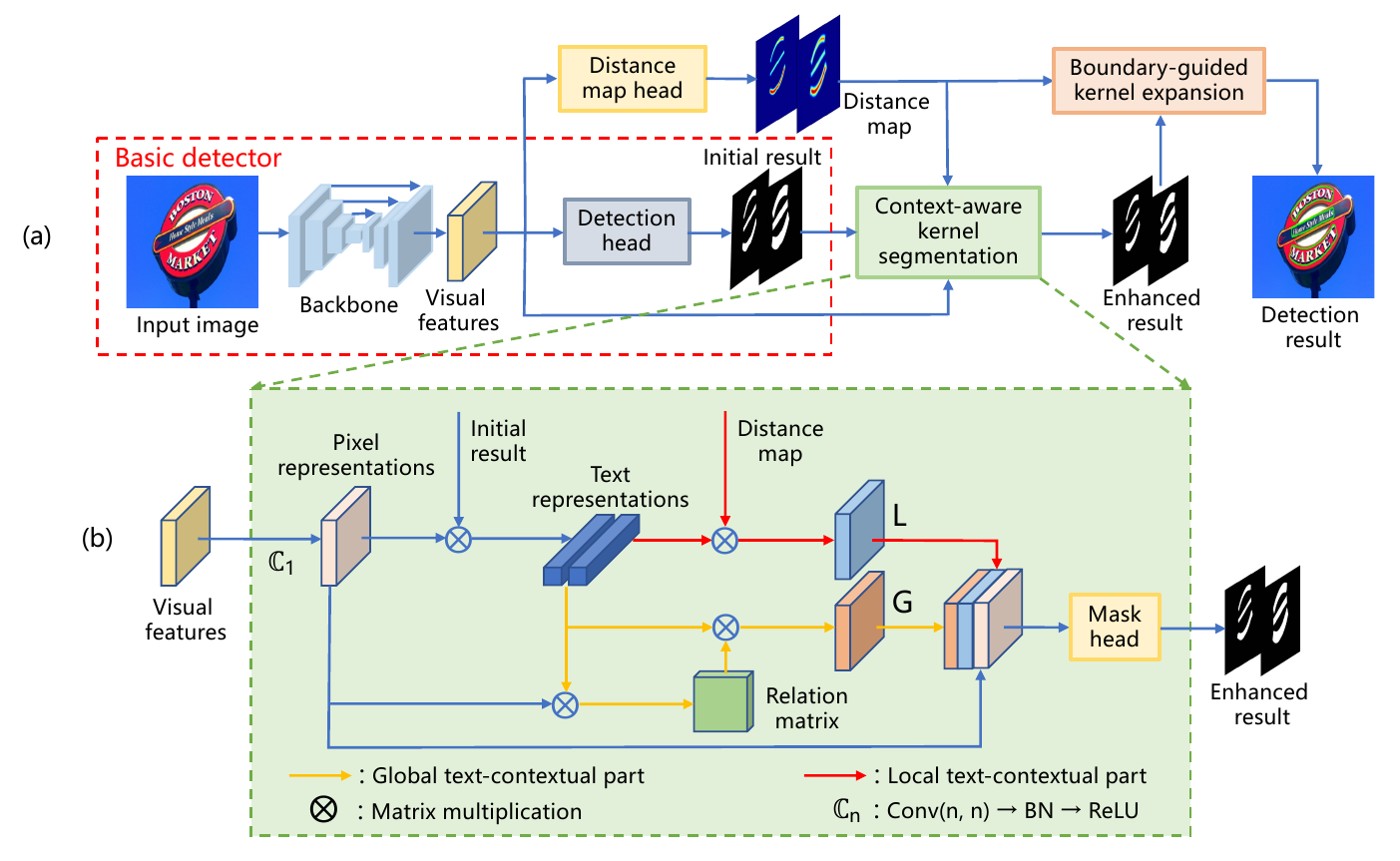}
	\caption{Detailed architecture of the proposed CBN. The context-aware module can enhance the initial text kernel segmentation result using global and local contexts. The boundary-guided module expands the enhanced text kernel with the predicted distance map. (b) illustrates the architecture of the context-aware text kernel segmentation module. ``$L$" and ``$G$" represent the local and global contexts respectively.}
	\label{fig:overview}
\end{figure*}

In this paper, we propose a \textbf{C}ontext-aware and \textbf{B}oundary-guided \textbf{N}etwork (CBN) to remedy the above problems as shown in Fig.~\ref{fig:overview}. Specifically, a segmentation-based method is adopted as the basic text detector, which predicts the initial segmentation result. Based on the output of the basic detector, we design a context-aware module to learn the relationship between each pixel, in which the relationship between the pixels and their corresponding text instances, and the relationship with all text instances are encoded into local and global contexts respectively. With the explicit introduction of global and local contexts, the initial text kernel segmentation result is enhanced as shown in Fig.~\ref{subfig:cbn}. Finally, a boundary-guided module is introduced to expand the enhanced text kernel adaptively with a learnable distance map, which describes the nearest distance between each positive pixel and its corresponding text boundary. As this module only uses the pixels on the contour to predict the position of the boundary as shown in Fig.~\ref{subfig:bgpost}, the expansion process achieves a good balance between speed and performance.

In summary, our contributions are as follows:
\begin{enumerate}
	\item We propose a context-aware module for text kernel segmentation, in which the relationship between pixels is encoded as global and local contexts. Benefiting from the enhancement of feature representations by contexts, the initial text kernel segmentation result is improved.
	\item A boundary-guided module is proposed to expand the text kernel adaptively with only the pixels on the contour, which solves the contradictory requirement between segmentation accuracy and inference efficiency in output map resolution.
	\item The proposed CBN is a lightweight plug-and-play module, which can be readily plugged into existing segmentation-based text detectors. In particular, when equipped with CBN, the basic detector with a lightweight backbone can achieve state-of-the-art performance on several scene text detection benchmarks, while still keeping a competitive inference speed.
\end{enumerate}

\section{Related Work}
Recent scene text detection methods can be roughly divided into two categories: regression-based and segmentation-based methods. Here, we first introduce two different kinds of methods for scene text detection. Then we present some practical applications of scene text detection.

\textbf{Regression-based methods}~\cite{liao2017textboxes,liu2017deep,zhou2017east,ma2018arbitrary,liao2018textboxes++} take scene texts as general objects and regress the bounding boxes of text instances via general detection frameworks~\cite{2017Faster,liu2016ssd,he2017mask}. TextBoxes~\cite{liao2017textboxes} modified the anchor scales and shape of convolution kernels based on SSD~\cite{liu2016ssd} to handle the irregular text with various aspect ratios. Based on Faster R-CNN~\cite{2017Faster}, RRPN~\cite{ma2018arbitrary} introduced rotated region proposals to detect arbitrary-oriented scene texts. EAST~\cite{zhou2017east} is an anchor-free method, which used FCN~\cite{long2015fully} to directly predict pixel-level score map, rotation angle, and text boxes. DGGR~\cite{zhang2020deep} first used a graph convolutional network to learn and infer the linkage relationships of text components and then grouped them into text instances by linking merging. Although regression-based methods usually enjoy simple post-processing steps, most of them rely on complex anchor design and fail to deal with the text instance with arbitrary shape.

To solve this problem, many modules are elaborately designed to model the text boundary with better representation. FCENet~\cite{zhu2021fourier} modeled text instance contours with Fourier curve, while ABCNet~\cite{liu2020abcnet} and ABCNet-v2~\cite{liu2021abcnet} used Bezier curve to fit arbitrary closed contours with progressive approximation. TextBPN~\cite{zhang2021adaptive} and TextBPN++~\cite{zhang2022arbitrary} proposed boundary proposal module to generate coarse boundary proposals, then iteratively refined the coarse boundary points with the assistance of graph convolutional network or Transformer network. Although these methods deal with the text instance with arbitrary shapes, the performances of these methods are largely affected by the number of control points and the resolution of feature maps.

\textbf{Segmentation-based methods}~\cite{zhang2016multi,yao2016scene,lyu2018mask,deng2018pixellink,long2018textsnake,wang2019shape,xie2019scene,xu2019textfield,wang2019efficient,liao2020real,zhu2021textmountain} mainly focus on pixel-level feature representations, which usually first segment text kernels and expand them with post-processing algorithms. For text kernel segmentation, most existing methods adopt the fully convolutional network~\cite{long2015fully} to classify each pixel independently, where the shrunk text region is regarded as the positive sample region. The main difference includes the number of text kernels and the shrunk ratios when shrinking text polygons. For example, PSENet~\cite{wang2019shape} obtained different kernel scales by shrinking the original text instance sequentially. DB~\cite{liao2020real} generated a positive area by shrinking the original polygon with a small shrink ratio. For text kernel expansion, the existing methods can be divided into two categories: fixed value expansion~\cite{lyu2018mask,xie2019scene,liao2020real} and pixel aggregation~\cite{deng2018pixellink,long2018textsnake,xu2019textfield,wang2019efficient,wang2019shape,zhu2021textmountain}. For fixed value expansion, these methods usually extend the text polygons via some certain preset values. For example, DB~\cite{liao2020real} rebuilt the complete text contour by expanding the predicted shrunk text contour outward by a fixed offset value. Mask TextSpotter~\cite{lyu2018mask} and SPCNet~\cite{xie2019scene} performed the bilinear interpolation operation on the mask according to the aspect ratio of the RoI. For pixel aggregation, these methods usually aggregate all text pixels to different text instances by using some rules. For example, PAN~\cite{wang2019efficient} considered the text instances as clusters and the kernels of text instances are cluster centers and used the idea of clustering to guide the pixels in the text regions to their corresponding kernels. TextMountain~\cite{zhu2021textmountain} grouped text instance pixels to text lines by using the grads of TCBP, which planned a road to the mountaintop for each pixel on the mountain foot at the aggregation stage.

Although these methods are more robust to shape changes compared to regression-based methods, there are still some problems to be solved. Firstly, existing segmentation-based methods regard the text kernel prediction as a purely pixel-level segmentation task, which lacks the representation and learning between text instances. In addition, post-processing algorithms usually use a fixed offset or aggregate all positive pixels to rebuild the complete text contours. This undermines the accuracy and efficiency of the method. Inspired by attention-based methods~\cite{wang2018non,2019ACFNet,yuan2020object} in other computer vision tasks, we propose an efficient and lightweight plugin for segmentation-based detectors to learn the relationship between each pixel and design an accurate and efficient expansion process with a learnable distance map.

\textbf{Practical applications of scene text detection.} Scene text detection is an essential component in scene text reading, which is of great importance in both academia and industry due to its wide practical applications. There are numerous applications with texts found in scene images and video frames, such as indexing of multimedia resources, multimodal information understanding, license plate recognition for vehicle identification, traffic sign detection, \emph{etc}. These numerous applications can be divided into two types: server-based applications and client applications on edge devices. Both require consideration of the model complexity and inference speed. Segmentation-based methods~\cite{wang2019shape,wang2019efficient,liao2020real} have wide applications in the industry due to their speed and accuracy. However, the speed and performance are still limited by model structure and post-processing for the complete text reconstruction. Motivated by this, we propose a lightweight plug-and-play for segmentation-based detectors, which can improve the performance of basic detectors and keep a competitive inference speed.

\section{Methodology}
In this section, we first introduce the overall pipeline of the proposed \textbf{C}ontext-aware and \textbf{B}oundary-guided \textbf{N}etwork (CBN). Then, we present the details of context-aware text kernel segmentation and boundary-guided text kernel expansion respectively. 

\subsection{Overall Architecture}
The overview of our proposed architecture is illustrated in Fig.~\ref{fig:overview}. Firstly, a segmentation-based text detector is used as the basic detector to predict the initial result, which consists of a backbone and a detection head. The backbone is utilized to extract the visual features of the input image, which concatenates low-level texture features with high-level semantic features. The extracted visual features are sent to the detection head for specific detection tasks. Then, based on the initial segmentation result, we design a context-aware module to learn the relationship between each pixel, and this relationship is encoded into global and local contexts through the links between the pixel and text instances. After that, the initial segmentation result is enhanced by the fused global and local contexts. Finally, based on the enhanced segmentation result, a boundary-guided module is used to expand the enhanced text kernels adaptively, in which a learnable distance map guides the pixels on the contour to restore the text boundary quickly and accurately.

\subsection{Context-aware Text Kernel Segmentation}
Most segmentation-based text detection methods consider each pixel classification as independent from each other when designing network, which lacks learning of the relationship between each pixel. Therefore, we design a context-aware text kernel segmentation module for learning the relationship between each pixel and enhancing the initial text kernel segmentation result. As shown in Fig.~\ref{fig:overview}(b), the context-aware module consists of global and local text-contextual parts.

\subsubsection{Global Text-contextual}
For the global text-contextual part, we firstly get the pixel representations $\textbf{P} \in \mathbb{R}^{C \times H \times W}$ from the visual features by using the function $\mathbb{C}_{n}(\cdot)$, where $C$, $H$ and $W$ denote the channel, height and width of the pixel representations respectively. The function $\mathbb{C}_n(\cdot)$ refers to $Conv(n \times n) \to BN \to ReLU$ and here $n = 1$. Then, we use $k$th initial segmentation map $\textbf{S}_k \in \mathbb{R}^{H \times W}$ as the weight map for $\textbf{P} \in \mathbb{R}^{C \times H \times W}$ and aggregate the weighted pixel representations on the spatial dimension, forming the $k$th text representation ${\textbf{T}}_k$:

\begin{equation}
	\label{equ:fi}
	\textbf{T}_k = \textbf{P} \times \textbf{S}_k.
\end{equation}
During multiplication, we need to reshape the dimension of tensor ${\textbf{P}}$ and ${\textbf{S}}_k$ to ($C$, $H \times W$) and ($H\times W$, 1) respectively.

To obtain the global relationship between pixel representations and all text instances, we compute the global relation matrix $\textbf{M}$ as below:
\begin{equation}
	\label{equ:grelation}
	\textbf{M}_{ki} =\frac{e^{\kappa({\textbf{T}}_{k}, {\textbf{P}}_{i})}}{\sum_{j=1}^{K}e^{\kappa({\textbf{T}}_{j}, {\textbf{P}}_{i})}}.
\end{equation}
Here, ${\textbf{P}}_i \in \mathbb{R}^{C \times 1}$ is the $i$th pixel representation vector. $\kappa({\textbf{T}}, {\textbf{P}}) = \phi({\textbf{T}})^\top \times \psi({\textbf{P}})$ is the unnormalized relation function, so we use the softmax function to normalize it. $\phi(\cdot)$ and $\psi(\cdot)$ are two transformation functions implemented by $Conv1d(n)$ and here $n = 1$. The dimension of tensor $\textbf{M}$ is ($K$, $H \times W$). $K$ is the number of initial segmentation results.

Finally, the global text-contextual representation ${\textbf{G}}_i$ for pixel $i$ is computed according to the following:
\begin{equation}
	\label{equ:global}
	{\textbf{G}}_{i} = \rho(\sum_{k=1}^{K}\textbf{M}_{ki}\delta({\textbf{T}}_{k})),
\end{equation}
and this equation can also be rewritten using matrix multiplication: 
\begin{equation}
	\label{equ:matglobal}
	\textbf{G} = \rho(\delta(\textbf{T}) \times \textbf{M}).
\end{equation}
$\rho(\cdot)$ and $\delta(\cdot)$ are implemented by $Conv1d(n)$ and $n=1$. The global text-contextual representation $\textbf{G}$ has the same size as the pixel representations $\textbf{P} \in \mathbb{R}^{C \times H \times W}$. The global text-contextual $\textbf{G}$ is obtained by weighting the text representations $\textbf{T} \in \mathbb{R}^{C \times K}$ with the relation matrix $\textbf{M}$, which contains the relationship between the pixel representations and the representations of all text instances. Therefore, $\textbf{G}$ can be considered as global contextual information.

\begin{figure*}[t!]
	\centering
	\includegraphics[width=\linewidth]{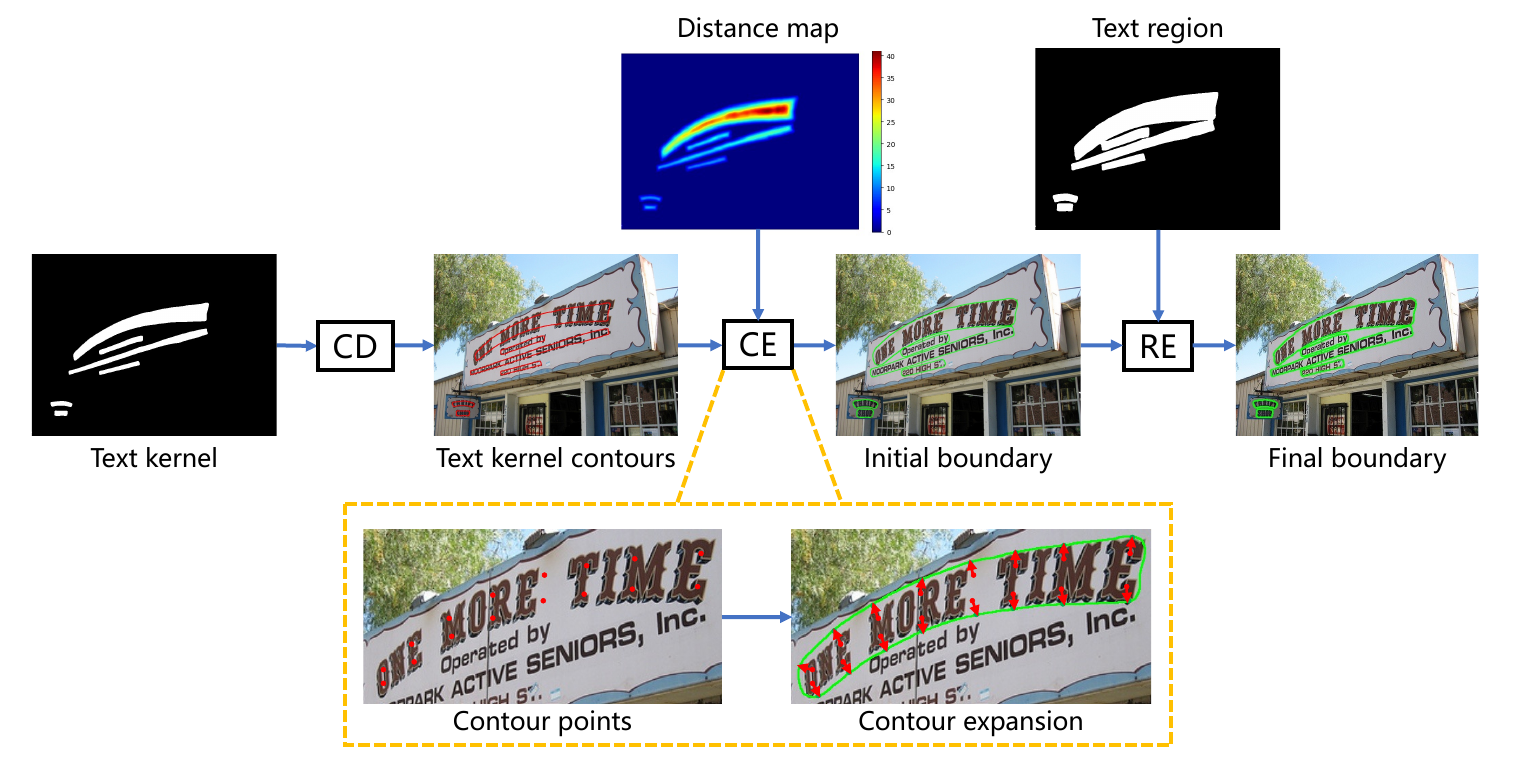}
	\caption{The procedure of boundary-guided kernel expansion algorithm. ``CD" refers to the contour detection. ``CE" represents the contour expansion. ``RE" refers to refining the initial boundary. The yellow dashed box shows the details of contour expansion.}
	\label{fig:BG}
\end{figure*}

\subsubsection{Local Text-contextual}

The global text-contextual module establishes the contextual relationships from the perspective of each pixel representation and all text representations. To construct a complementary relationship with the global text-contextual module, we propose a local text-contextual module to represent the relationship between each text pixel and its corresponding text instance. This module shares the same text representation ${\textbf{T}}_{k}$ with the global text-contextual module. The local text-contextual computes the context for pixel $i$ by considering the relations:

\begin{equation}
	\label{equ:local}
	{\textbf{L}}_{i} = \rho(\sum_{k=1}^{K}Sigmoid({\textbf{D}}_{ki})\delta({\textbf{T}}_{k})).
\end{equation}
Here, ${\textbf{T}}_{k}$ is calculated by Eq.~\ref{equ:fi}. ${\textbf{D}}_{ki}$ is obtained by using a distance map head on visual features (see Fig.~\ref{fig:overview}(b)), which encodes the relationship between pixels within a text instance and provides important adaptive values for boundary-guided text kernel expansion. From the perspective of matrix multiplication, Eq.~\ref{equ:local} can also be rewritten as: 

\begin{equation}
	\label{equ:matlocal}
	{\textbf{L}} = \rho(\delta({\textbf{T}}) \times Sigmoid({\textbf{D}})).
\end{equation}
Here the distance map $\textbf{D} \in \mathbb{R}^{K \times H \times W}$ is reshaped to ($K$, $H \times W$). The local text-contextual representation $\textbf{L}$ has the same size as the pixel representations $\textbf{P}\in \mathbb{R}^{C \times H \times W}$. Since $\textbf{L}$ is obtained by weighting the text representations $\textbf{T}$ with the learned distance map $\textbf{D}$, which represents the relationship between the pixels and their corresponding text instances. $\textbf{L}$ can be considered as local contextual information.

Given the global context ${\textbf{G}} \in \mathbb{R}^{C \times H \times W}$, the local context ${\textbf{L}} \in \mathbb{R}^{C \times H \times W}$, and the pixel representations $\textbf{P} \in \mathbb{R}^{C \times H \times W}$, the fused features ${\textbf{F}}$ can be obtained by the following equation:

\begin{equation}
	\label{equ:fused}
	{\textbf{F}} = [{\textbf{G}},~ {\textbf{L}},~ {\textbf{P}}],
\end{equation}
where $[\cdot, \cdot, \cdot]$ denotes concatenation. Next, the fused features ${\textbf{F}}$ pass through a mask head to produce the enhanced segmentation result $\textbf{S}'$. The mask head consists of a $\mathbb{C}_3(\cdot)$ operation and a $1\times1$ convolution layer. 

Considering the extreme imbalance of text and non-text pixels, we employ the dice loss~\cite{milletari2016v} for supervising $S'$. The loss function $\mathcal{L}_{seg}$ can be formulated as:
\begin{equation}
	\label{equ:diceloss}
	\mathcal{L}_{seg}(\textbf{S}', \hat{\textbf{S}'}) = 1 - \frac{2\sum_i(\textbf{S}'_i \hat{\textbf{S}'_i})}{\sum_i(\textbf{S}'_i)^2 + \sum_i(\hat{\textbf{S}'_i})^2},
\end{equation}
where $\textbf{S}'_i$ and $\hat{\textbf{S}'_i}$ represent the predicted value and ground truth value of pixel $i$.

\subsection{Boundary-guided Text Kernel Expansion}
Existing kernel expansion methods rebuild the complete text contours by a non-adaptive expansion distance or pixel aggregation. These methods are difficult to achieve a satisfactory trade-off between speed and accuracy. Therefore, we propose the boundary-guided text kernel expansion to guide the text kernel contours towards the accurate text contours efficiently. We use a learnable distance map $\textbf{D} \in \mathbb{R}^{K \times H \times W}$ to describe the distance from each pixel to the nearest boundary pixel, which is obtained by using a distance map head on visual features (see Fig.~\ref{fig:overview}(b)). For the label generation of the distance map, we calculate the minimum distance $\hat{\textbf{D}_{i}}$ between the text pixel $i$ and its corresponding boundary pixels:
\begin{equation}
	\label{equ:distlabel}
	\hat{\textbf{D}_{i}} = \min_{b \in \mathbb{B}} \left \{ \left \| p_b - p_i \right \|_2 \right \}, 
\end{equation}
where $\mathbb{B}$ is the set of boundary pixels of the text instance to which $i$ belongs. $p_b$ and $p_i$ are the coordinates of $b$ and $i$. Here, we utilize the distance transform algorithm in OpenCV to generate the distance map.

To learn the distance map, we adopt the distance ratio loss $\mathcal{L}_{dis}$ which can be defined as:
\begin{equation}
	\label{equ:ratioloss}
	\mathcal{L}_{dis}(\textbf{D}_i, \hat{\textbf{D}_i}) = \ln\frac{\max(\textbf{D}_{i}, \hat{\textbf{D}}_{i})}{\min(\textbf{D}_{i}, \hat{\textbf{D}}_{i})},
\end{equation}
where $\textbf{D}_{i}$ and $\hat{\textbf{D}}_{i}$ denote the predicted distance at pixel $i$ and its ground truth respectively, and pixel $i$ is located in the text region. $\textbf{D}_{i}$ is generated from a distance map head, which is implemented by a $1 \times 1$ convolution layer.

In the inference period, we use the predicted distance map to guide the contour points of the text kernel to expand to the text boundary as shown in Fig.~\ref{fig:BG}. The detailed post-processing steps are as follows: 

\begin{enumerate}[(1)]
	\item Detect the contour point set $Con_i$ of $i$th text kernel. 
	\item Obtain the expanded values corresponding to $Con_i$ on the distance map and average them, then utilize the Vatti clipping algorithm~\cite{vatti1992generic} to expand $Con_{i}$ to $ECon_{i}$.
	\item Generate the text instance map ${Te}_{i}$ through the expanded contour $ECon_{i}$, and then multiply ${Te}_{i}$ with the text region map $R$ to refine ${Te}_{i}$. 
	\item Find the contours of ${Te}_{i}$ to obtain the final boundary of $i$th text instance. As the computation of each text instance is independent, our post-processing can be implemented in parallel.
\end{enumerate}

Benefiting from the advantages of the learned distance map, our method can reduce the influence of the quality of text kernel segmentation on the final detection result, and only using contour points in the expansion process greatly reduces the amount of calculation. Moreover, since the text contours can be reconstructed by extending the text kernel contours outward directly, CBN enjoys simpler post-processing than non-real-time methods (such as~\cite{zhang2020deep,zhu2021fourier,dai2021progressive}), which makes our method run faster.

\subsection{Loss Function}
For optimizing our model, the overall loss function $\mathcal{L}$ can be formulated as:
\begin{align}
	\label{equ:loss}
	\mathcal{L} &= \mathcal{L}_{B} + \mathcal{L}_{CBN}, \\
	\label{equ:losscbn}
	\mathcal{L}_{CBN} &= \lambda_1\mathcal{L}_{seg} + \lambda_2\mathcal{L}_{dis},
\end{align}
where $\mathcal{L}_{B}$ is the loss of the basic detector, and $\mathcal{L}_{CBN}$ is the relevant loss of the proposed CBN. For $\mathcal{L}_{CBN}$ in Eq.~\ref{equ:losscbn}, $\mathcal{L}_{seg}$ (see Eq.~\ref{equ:diceloss}) represents the loss for the enhanced segmentation maps. $\mathcal{L}_{dis}$ (see Eq.~\ref{equ:ratioloss}) represents the loss for the distance maps. $\lambda_1$ and $\lambda_2$ are loss weights, which are set to 1 and 0.25 in
our experiments. Note that, we apply Online Hard Example Mining (OHEM)~\cite{shrivastava2016training} to ignore simple non-text pixels while calculating $\mathcal{L}_{CBN}$.

\section{Experiments}

\subsection{Datasets}
For the experimental part, we employ six commonly used public datasets as follows:

\textbf{SynthText}~\cite{gupta2016synthetic} is a large-scale synthetic dataset, consisting of more than 800K synthetic images. This dataset is used to pretrain our model.

\textbf{CTW1500}~\cite{yuliang2017detecting} is a challenging dataset for curve text, which consists of 1,000 training images and 500 testing images. The text instances are annotated in line-level.

\textbf{Total-Text} ~\cite{ch2017total} is a curve text detection dataset, which includes 1,255 training images and 300 testing images. This dataset contains horizontal, multi-oriented, and curve text lines. They are labeled in word-level.

\textbf{MSRA-TD500}~\cite{yao2012detecting} is a diverse dataset that includes multi-lingual, multi-oriented, and long text lines. It contains 300 training images and 200 testing images with line-level annotation. Since the training set is rather small, we follow the previous works~\cite{zhou2017east,liao2020real,wang2019efficient} to use extra 400 images from HUST-TR400~\cite{yao2014unified} as training data.

\textbf{ICDAR 2015} is a commonly used dataset for text detection, which consists of 1000 training images and 500 testing images. The text instances are annotated by quadrangles in word-level.

\textbf{MTWI}~\cite{he2018icpr2018} is a large-scale dataset for Chinese and English web text reading, which contains some challenging cases, such as complex layout, watermarks, and small text. There are 10,000 training images and 10,000 testing images, and all text instances are labeled in line-level.

\textbf{ICDAR-MLT19}~\cite{nayef2019icdar2019} is a multi-language dataset, which includes 10 languages representing 7 different scripts. The languages include Chinese, Japanese, Korean, English, French, Arabic, Italian, German, Bangla, and Hindi (Devanagari). It consists of 10,000 training images, 2,000 validation images, and 10,000 testing images. We use the training set in the finetune period.

\textbf{COCO-Text}~\cite{veit2016coco} is one of the challenges of ICDAR 2017 Robust Reading Competition, which consists of 43,686 images for training, 10,000 images for validation, and another 10,000 for test. Although text instances in this dataset are oriented, each text is labeled in the form of axis-aligned bounding box.

\begin{table*}[t!]
	\centering
	\caption{The results of different context-aware modules on CTW1500. ``1s" and ``4s" mean the width and height of the output map is 1/1 and 1/4 of the input image. ``CAM" means our proposed context-aware module. ``G" and ``L" represent global and local respectively. ``Time" denotes the time consumption of the model inference.}
	\resizebox{.9\linewidth}{!}{
		\begin{tabular}{c|c|c|c|c|c|c}
			\hline
			\textbf{Method}   & \textbf{Precision} &\textbf{Recall} &\textbf{F-measure} &\textbf{\textbf{Time} (ms)} &\textbf{GFLOPs(G)} &\textbf{Params(M)}\\
			\hline
			\hline
			PAN-4s (baseline)   & 86.2 & 80.9 &83.5 &25.4  &43.7214 &12.2460\\ 
			+G-CAM   & \textbf{88.1} & 80.5 &84.1   &25.8  &43.9918 &12.2566 \\
			+L-CAM  &  87.2 & 81.1  &84.1  &26.1 &43.9971  & 12.2568 \\
			+GL-CAM   & \textbf{88.1} & \textbf{81.6} & \textbf{84.7}  &26.2 &44.0040 &12.2570\\
			\hline
			PAN-1s (baseline)    & 86.5 & 81.1 &83.7  &27.3  &43.7214 &12.2460\\
			+G-CAM   & \textbf{89.1} & 80.1 &84.4  &27.6  &43.9918 &12.2566\\
			+L-CAM    & 88.1 & 81.2  &84.5   &27.8 &43.9971 & 12.2568 \\
			+GL-CAM    & 89.0 &\textbf{ 81.9} & \textbf{85.3}    &27.9  &44.0040 & 12.2570\\
			\hline  
		\end{tabular}
	}
	\label{tab:ablation_context}
\end{table*}

\begin{table*}[t!]
	\centering
	\caption{Experiment results of other popular context-aware modules on CTW1500. ``4s" means the width and height of the output map is 1/4 of the input image. ``GL-CAM" means our proposed context-aware module. Note that all models are trained from scratch without using any external training data.}
	\resizebox{.85\linewidth}{!}{
		\begin{tabular}{c|c|c|c|c|c|c}
			\hline
			\textbf{Method}  & \textbf{Precision} &\textbf{Recall}&\textbf{F-measure} &\textbf{FPS} &\textbf{GFLOPs(G)} &\textbf{Params(M)} \\
			\hline
			\hline
			PAN-4s (baseline)  &85.1 &79.1 &82.0  &\textbf{30.8} &43.7214 &12.2460 \\ 
			+NL~\cite{wang2018non}  &85.7 &78.8 &82.1 &13.2 &43.7233 &12.2535 \\
			+GCN~\cite{cao2019gcnet}  &85.5  &79.8 &82.6  &30.4 &43.7215 &12.2534 \\
			+PCAA~\cite{liu2022partial}  &85.5 &\textbf{80.4} &82.8  &30.2 &43.7225 &12.2538 \\
			+GL-CAM (Ours) &\textbf{87.2} &79.8 &\textbf{83.3} &30.5 &44.0040 &12.2570 \\ 
			\hline
		\end{tabular}
	}
	\label{tab:ablation_context2}
\end{table*}

\subsection{Implementation Details}
The backbone of our method is pretrained on ImageNet~\cite{deng2009imagenet}. All the networks are optimized by the Adam~\cite{kingma2014adam} optimizer. We pre-train the model on SynthText~\cite{gupta2016synthetic} for 5 epochs with a fixed learning rate of $1\times10^{-3}$. Then, we finetune the models on each corresponding real-world dataset. Notably, in some experiments, we also fine-tune our models on the ICDAR-MLT19 model for more comprehensive comparisons. In fine-tuning stage, we train our models with batch size 16 on 2 GPUs for 600 epochs and use a ``poly" learning rate strategy where the initial learning rate and power are set to $1 \times10^{-3}$ and $0.9$ respectively. The data augmentation for the training images includes random horizontal flip, random scale, random crop, and random rotation. In the inference period, we keep the aspect ratio of the test images and resize the input images on each dataset. All experiments are conducted with the deep learning framework PyTorch~\cite{paszke2017automatic}, and on a workstation with two Tesla P40 GPUs, a 2.10GHz Intel(R) Xeon(R) CPU E5-2683-v4 CPU, and 56G RAM.

Notably, there are two datasets to pretrain models, ``Synth" donates SynthText, and ``MLT" donates ICDAR-MLT19. ``P", ``R", and ``F" represent precision, recall, and F-measure respectively. For a fair comparison of detection speed, the speed is evaluated in the same way as the basic detectors (such as PSENet~\cite{wang2019shape}, PAN~\cite{wang2019efficient}, and DBNet~\cite{liao2020real}).

\subsection{Ablation Studies} 
To analyze our proposed CBN in depth, we conduct a series of ablation studies on curve and multi-lingual datasets. If not specified, all the experiments are performed on the same hardware environment.

\subsubsection{Effectiveness of Context-aware Module}
This module aims to enhance the segmentation results, which can learn the relationship between each pixel and each text region and augment the representation of each text pixel. Here, we use PAN-1s and PAN-4s as our baseline models and add different context-aware modules to them. To ensure a fair comparison, we all use Pixel Aggregation (PA) mentioned in PAN~\cite{wang2019efficient} as post-processing.

\begin{figure}[t!]
	\centering					
	\begin{subfigure}[b]{.233\textwidth}
		\centering
		\includegraphics[width=\textwidth]{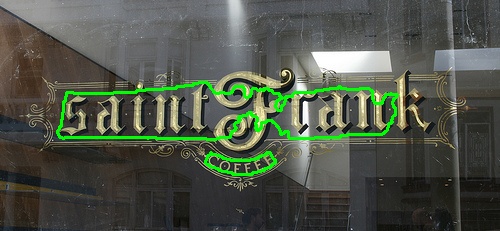}
		\vspace{.2pt}
		\includegraphics[width=\textwidth]{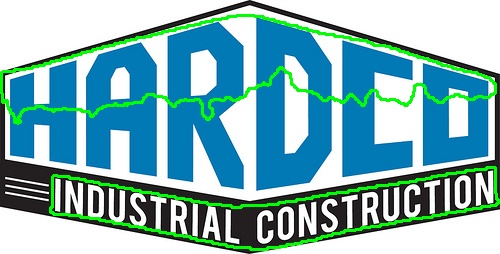}
		\caption{Without GL-CAM}
		\label{subfig:withoutgl}
	\end{subfigure}
	\begin{subfigure}[b]{.233\textwidth}
		\centering
		\includegraphics[width=\linewidth]{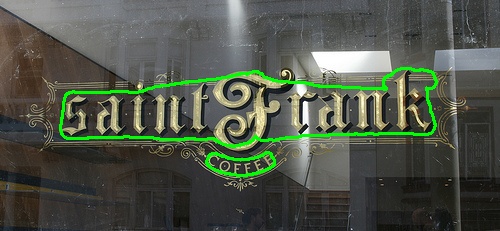}
		\vspace{.2pt}
		\includegraphics[width=\linewidth]{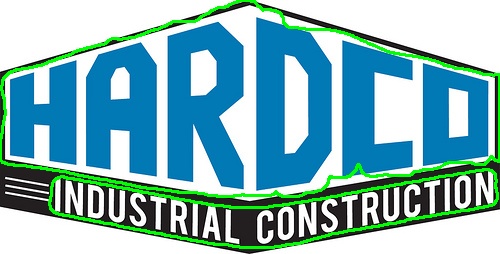}
		\caption{With GL-CAM}
		\label{subfig:withgl}
	\end{subfigure}
	\caption{Visualization results of the basic detector with and without the GL-CAM module. ``GL-CAM" means our proposed global and local context-aware module.}
	\label{fig:visualContext}
\end{figure}

As shown in Tab.~\ref{tab:ablation_context}, the PAN with different context-aware modules all outperform the baseline model. Adding a global or local context-aware module improves the performance at 1s and 4s output map scales, which demonstrates that the text contextual information can enhance the segmentation maps. Meanwhile, the F-measure on ``1s" and ``4s" can further be boosted to 85.3\% and 84.7\% when fusing the global and local context-aware modules. This indicates that the fused context-aware module can bring complementary gains to the detector. Since all the context-aware modules in Tab.~\ref{tab:ablation_context} only increase a few parameters and GFLOPs on the base detector, the time consumption of the model part is similar. We also visualize the detection results of the basic detector with and without the GL-CAM module in Fig.~\ref{fig:visualContext}, which shows that our context-aware module can refine the segmentation results and improve the detection result.

To fully validate the effectiveness of our proposed GL-CAM, we also make a comparison with other public context-aware modules on CTW1500, such as NL~\cite{wang2018non}, GCNet~\cite{cao2019gcnet} and PCAA~\cite{liu2022partial}. As shown in Tab.~\ref{tab:ablation_context2}, the number of model parameters is very close, but the PAN baseline equipped with our GL-CAM can keep high speed and achieve the highest F-measure of 83.3\%, which outperforms all other context-aware modules by 1.2\%, 0.7\% and 0.5\% on CTW1500 dataset respectively. This indicates that our GL-CAM is more effective than other popular context-aware modules.

\begin{table*}[t!]
	\centering
	\caption{The results of different post-processing methods on CTW1500. ``CAM" means the context-aware module. ``PA" means pixel aggregation post-processing. ``DB'' means the post-processing of DBNet. ``BG" means our boundary-guided kernel expansion post-processing.}
	\resizebox{.85\linewidth}{!}{
		\begin{tabular}{c|c|cc|c|c|c|c}
			\hline
			\multirow{2}{*}{\textbf{Method}} & \multirow{2}{*}{\textbf{Post-processing}}  & \multicolumn{2}{c|}{\textbf{\makecell[c]{Time consumption \\(ms)}}} & \multirow{2}{*}{\textbf{Precision}} &  \multirow{2}{*}{\textbf{Recall}} & \multirow{2}{*}{\textbf{F-measure}} & \multirow{2}{*}{\textbf{FPS}} \\
			\cline{3-4}
			& &\multicolumn{1}{c|}{\textbf{Model part}} & \textbf{Post part}& & &\\
			\hline
			\hline
			\multirow{3}{*}{\makecell[c]{PAN-4s \\ +GL-CAM}}  & PA   & \multicolumn{1}{c|}{26.2} &7.6  &\textbf{89.3}  &82.0 &85.5 &29.5 \\
			& DB   & \multicolumn{1}{c|}{26.2} & 5.8 &79.1   & \textbf{83.4} & 81.2 &\textbf{31.3} \\ 
			& BG  & \multicolumn{1}{c|}{26.2} & 6.4 &88.5 &83.1 &\textbf{85.7} &30.7  \\
			\hline
			\multirow{3}{*}{\makecell[c]{PAN-1s \\ +GL-CAM}}  & PA   & \multicolumn{1}{c|}{26.2} & 110.7 &\textbf{89.5}  &82.6 &85.9 &7.3 \\
			& DB   & \multicolumn{1}{c|}{26.2} &8.5 & 80.9   & \textbf{84.1 } & 82.4  &\textbf{28.8}  \\
			& BG  & \multicolumn{1}{c|}{26.2} &9.2 &89.3 &82.9  &\textbf{86.0}  &28.2  \\
			
			\hline
		\end{tabular}
	}
	\label{tab:ablation_bg}
\end{table*}

\begin{figure*}[t!]
	\centering
	\begin{subfigure}[b]{.3\textwidth}
		\centering
		\includegraphics[width=\textwidth]{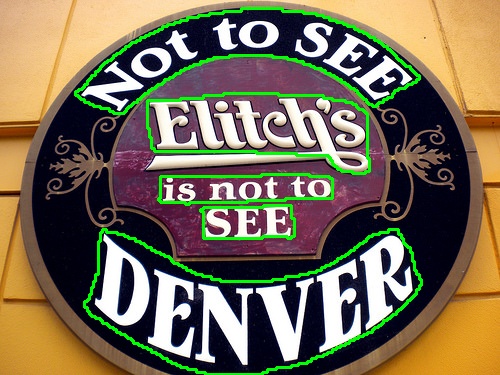}
		\vspace{.2pt}
		\includegraphics[width=\textwidth]{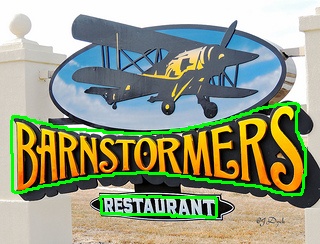}
		\caption{PA}
		\label{subfig:pa}
	\end{subfigure}
	\begin{subfigure}[b]{.3\textwidth}
		\centering
		\includegraphics[width=\textwidth]{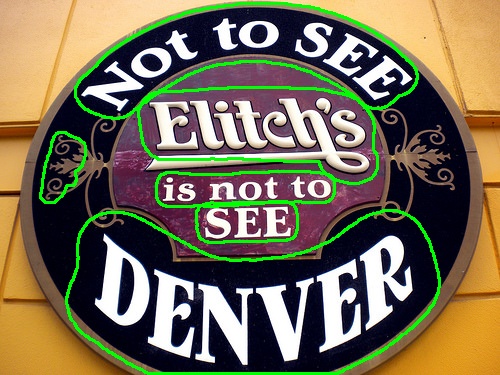}
		\vspace{.2pt}
		\includegraphics[width=\textwidth]{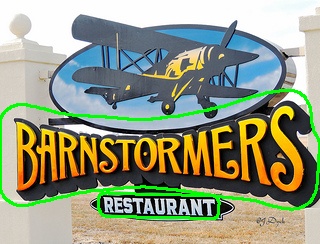}
		\caption{DB}
		\label{subfig:db}
	\end{subfigure}
	\begin{subfigure}[b]{.3\textwidth}
		\centering
		\includegraphics[width=\textwidth]{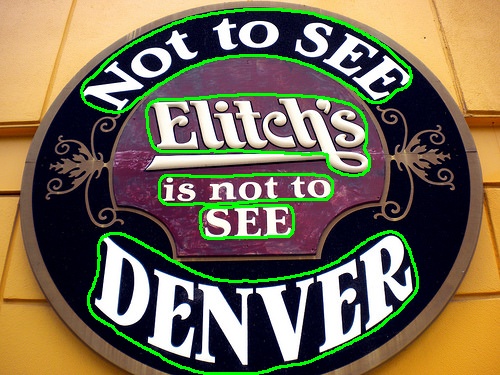}
		\vspace{.2pt}
		\includegraphics[width=\textwidth]{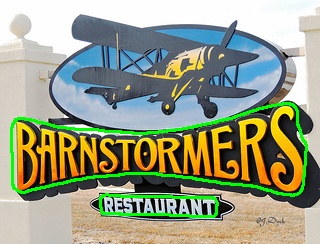}
		\caption{BG}
		\label{subfig:bg}
	\end{subfigure}
	\caption{Visualization results of PAN equipped with GL-CAM by using different post-processing. ``PA", ``DB", and ``BG" represent pixel aggregation, post-processing of DBNet, and our boundary-guided kernel expansion respectively.}
	\label{fig:visualBG}
\end{figure*}

\begin{table}[h!]
	\centering
	\caption{The effect of the number of control points on TextBPN++ and our method on CTW1500. ``\ding{55}" means that there are no control points in our method. ``P", ``R", and ``F" represent precision, recall, and F-measure respectively.}
	\resizebox{\linewidth}{!}{
		\begin{tabular}{c|c|c|c|c|c}
			\hline			
			\textbf{Method} &\textbf{Control points} &\textbf{P(\%)} & \textbf{R(\%)} & \textbf{F(\%)} & \textbf{FPS} \\
			\hline
			\hline
			\multirow{3}{*}{TextBPN++~\cite{zhang2022arbitrary}} &8  &85.2 &82.4 &83.8 &10.0  \\
			 & 20   &87.3 &\textbf{83.8} &85.5 &8.6\\
			  & 32   &87.4 &82.6 &84.9 &9.8  \\
			 \hline
			Ours  & \ding{55} &\textbf{89.0 }&83.0  &\textbf{85.9}  &\textbf{17.1}   \\
			\hline
		\end{tabular}
	}
	\label{tab:vstextbpcp}
\end{table}

\begin{table}[h!]
	\centering
	\caption{The effect of different feature resolutions on TextBPN++ and our method on CTW1500. ``1s" and ``4s" mean the width and height of the output feature map is 1/1 and 1/4 of the testing image.}
	\resizebox{\linewidth}{!}{
		\begin{tabular}{c|c|c|c|c|c}
			\hline			
			\textbf{Method} &\textbf{Scale}  &\textbf{Precision} & \textbf{Recall} & \textbf{F-measure} & \textbf{FPS}  \\
			\hline
			\hline
			TextBPN++~\cite{zhang2022arbitrary} &4s   &87.3 &80.9 &84.0 &13.2 \\
			Ours  &4s&88.2 &82.8 &85.4 &\textbf{18.5}   \\
			\hline
			TextBPN++~\cite{zhang2022arbitrary} &1s     &87.3 &\textbf{83.8} &85.5 &8.6 \\
			Ours &1s  &\textbf{89.0 }&83.0  &\textbf{85.9}  &17.1   \\
			\hline
		\end{tabular}
		
	}
	\label{tab:vstextbps}
\end{table}

\subsubsection{Effectiveness of Boundary-guided Module}
Boundary-guided text kernel expansion is a learnable post-processing algorithm for reconstructing the complete text instances, which can guide the text kernel contours towards the complete text contours by a learned distance map. To investigate the effectiveness of our proposed boundary-guided module, we use the same model, basic detector PAN equipped with our proposed global and local context-aware module (GL-CAM), and employ different post-processing methods for this model.

As shown in~Tab.~\ref{tab:ablation_bg}, when the post-processing is performed on 1s feature maps, the time consumption of PA post-processing is 101.5 ms slower (110.7 ms \emph{vs} 9.2 ms) than BG post-processing. This is because PA post-processing needs to aggregate all positive pixels, which leads to excessive time consumption when performing on the high-resolution output map. On the contrary, BG post-processing only uses the pixels on the contour when reconstructing each text line, and achieves the F-measure of 86.0\%, which is 0.1\% better than PA post-processing. In addition, although DB and BG post-processing have almost the same time consumption, the performance of BG achieves 3.6\% gains in F-measure due to the learnable text kernel expansion. When the post-processing is performed on 1/4 scale feature maps, the speed of BG post-processing is boosted to 30.7 FPS while the performance is still very competitive. These experiments demonstrate the efficiency and performance of our proposed boundary-guided post-processing for rebuilding the complete text instances. As shown in Fig.~\ref{fig:visualBG}, the detection results with ``PA" and ``BG" post-processing are significantly better than the post-processing of DBNet. Although both ``PA" and ``BG" post-processing achieve good detection results, the detection speed of ``PA" post-processing is slower than others, especially on the full-size feature map.

\begin{table*}[t!]
	\centering
	\caption{Popular segmentation-based methods with \emph{vs} without CBN on CTW1500 and Total-Text. For a fair comparison, we use their official open-source repository code to reproduce them with their corresponding default training parameter settings. ``P", ``R", and ``F" represent precision, recall, and F-measure respectively.}
	\resizebox{\linewidth}{!}{
		\begin{tabular}{c|c|c|c|c|c|c|c|c|c|c|c|c|c}
			\hline
			\multirow{2}{*}{\textbf{Method}} & \multirow{2}{*}{\textbf{Backbone}} &\multirow{2}{*}{\textbf{CBN}} 
			&\multicolumn{5}{c|}{\textbf{CTW1500}} & \multicolumn{5}{c|}{\textbf{Total-Text}}
			& \multirow{2}{*}{\textbf{Params(M)}}\\
			\cline{4-13}
			& &  &\textbf{Short Side} &\textbf{ P(\%)} & \textbf{R(\%)} & \textbf{F(\%) }   & \textbf{FPS}  &\textbf{Short Side} &\textbf{ P(\%)} & \textbf{R(\%)} & \textbf{F(\%) } & \textbf{FPS} &\\
			\hline
			\hline
			\multirow{2}{*}{PSENet} &   \multirow{2}{*}{Res50} &\ding{55} &  \multirow{2}{*}{640}   & 83.0& 75.1&78.9  &10.7 
			&  \multirow{2}{*}{736} &87.6 &76.2 &81.5  &10.9  &24.1096\\
			
			&  &\ding{51} &  & 83.3 & 77.5& 80.3 &11.1
			& &86.9 &78.1 &82.3  &11.0 &24.1159 \\
			
			\hline
			\multirow{2}{*}{PAN} & \multirow{2}{*}{Res18} & \ding{55}   &  \multirow{2}{*}{640} &85.1 &79.1 &82.0  &30.8 
			& \multirow{2}{*}{640} &88.0 &\textbf{79.4} &83.5  &32.3 &12.2460 \\
			&  & \ding{51}   & &\textbf{87.2}  &79.8& \textbf{83.3}  &30.5
			& &\textbf{90.1} &79.1 &\textbf{84.5}  &31.1 &12.2570 \\

			\hline
			\multirow{2}{*}{DBNet} &\multirow{2}{*}{Res50}  &\ding{55} &  \multirow{2}{*}{736}  & 82.3 & 78.0 &80.1  &20.2
			& \multirow{2}{*}{736} & 82.5 &78.7 &80.5 & 20.1 &27.9833 \\
			& & \ding{51}  &  & 81.5&  \textbf{80.5}& 81.0    & 19.3
			& & 84.5 &78.3 &81.3  &18.9 &27.9871 \\
			\hline
		\end{tabular}
	}
	\label{tab:segmethods}
\end{table*}

\begin{table*}[t!]
	\centering
	\caption{CBN generalization experimental results with different backbones on CTW1500 and Total-Text. The basic detector is PAN.}
	\resizebox{.85\linewidth}{!}{
		\begin{tabular}{c|c|c|c|c|c|c|c}
			\hline
			\multirow{2}{*}{\textbf{Backbone}}  &\multirow{2}{*}{\textbf{CBN}} 
			&\multicolumn{3}{c|}{\textbf{CTW1500}}  &\multicolumn{3}{c}{\textbf{Total-Text}}  \\
			\cline{3-8}
			
			&  &\textbf{Precision} & \textbf{Recall}  & \textbf{F-measure}  &\textbf{Precision} & \textbf{Recall}  & \textbf{F-measure}   \\
			\hline
			MobileNet-v3~\cite{howard2019searching} &\ding{55} &84.2 &75.5 &79.6  &88.3 &75.7 &81.5  \\
			MobileNet-v3~\cite{howard2019searching} &\ding{51} &\textbf{88.7} &73.2 &80.2  &87.6 &78.1 &82.5  \\
			\hline
			Swin-Tiny~\cite{liu2021swin}  &\ding{55}&83.2 &71.7 &77.0  &81.2 &68.9 &74.6  \\
			Swin-Tiny~\cite{liu2021swin} &\ding{51}& 83.2 &72.8 &77.7  &84.3 &69.8 &76.3  \\
			\hline
			Res18  &\ding{55} &85.1 &79.1 &82.0   &88.0 &\textbf{79.4} &83.5 \\
			Res18  &\ding{51} &87.2  &\textbf{79.8} &\textbf{83.3}  &\textbf{90.1 }&79.1 &\textbf{84.5}  \\
			\hline
		\end{tabular}
	}
	\label{tab:cbn_backbone}
\end{table*}

\begin{table*}[t!]
	\centering
	\caption{Ablation experiments for our method with different backbones on CTW1500, Total-Text, and ICDAR-MLT19. ``P", ``R", and ``F" represent precision, recall, and F-measure respectively.}
	\resizebox{.8\linewidth}{!}{
		\begin{tabular}{c|c|c|c|c|c|c|c|c|c}
			\hline
			\multirow{2}{*}{\textbf{Backbone}} 
			&\multicolumn{3}{c|}{\textbf{CTW1500}}  &\multicolumn{3}{c|}{\textbf{Total-Text}}  &\multicolumn{3}{c}{\textbf{ICDAR-MLT19}} \\
			\cline{2-10}
			
			&  \textbf{P(\%)} & \textbf{R(\%)}  & \textbf{F(\%)} &\textbf{P(\%)} & \textbf{R(\%)} & \textbf{F(\%)}    &\textbf{P(\%)} & \textbf{R(\%)}  & \textbf{F(\%)} \\
			\hline
			Res18 &89.3 &\textbf{82.9}&\textbf{86.0} &89.3 &\textbf{85.2} &87.2  &81.3 &67.3 &73.6 \\
			Res50 & 89.0 &82.8 &85.8 &\textbf{91.7 }&83.1& 87.2  &82.6 &69.5 &75.5 \\
			Res50-DCN &\textbf{89.9} &82.2 &85.9 &91.1 &83.5 &87.2  &\textbf{84.8} &\textbf{69.7} &\textbf{76.6} \\
			\hline
		\end{tabular}
	}
	\label{tab:backbone}
\end{table*}

\subsubsection{Comparisons with a Regression-based Method}
The performances of most regression-based methods rely on the number of control points or parameter settings and the resolution of feature maps. To analyze these two different methods, we choose a SOTA regression-based detector (TextBPN++~\cite{zhang2022arbitrary}) and conduct ablation studies on CTW1500. Note that, to ensure a fair comparison, we all use ResNet50 as the backbone.

Firstly, since there are no control points for our method to represent text boundary proposals or instances, the performance of our method is not affected by the number of control points. As shown in Tab.~\ref{tab:vstextbpcp}, the F-measure of TextBPN++ goes down from 84.9\% to 83.8\% with the decrease of the control point number. In addition, the performance of TextBPN++ is largely affected by the resolution of feature maps. In Tab.~\ref{tab:vstextbps}, the F-measure of TextBPN++ is reduced by 1.5\% (85.5\% \emph{vs} 84.0\%) when using smaller scale feature maps, but our method only drops 0.5\%. Meanwhile, our method also outperforms TextBPN++ on speed. This indicates that our method can achieve a good balance between speed and performance when compared with TextBPN++.

\subsubsection{Generalization of CBN}
To better analyze the generalization of the proposed CBN, we use three popular segmentation-based methods (PSENet~\cite{wang2019shape}, PAN~\cite{wang2019efficient} and DBNet~\cite{liao2020real}) as basic detectors, and add the CBN to their official code repository. To avoid the influence of extra pre-training datasets (such as SynthText), we train them from scratch with their corresponding default parameter settings on CTW1500 and Total-Text datasets. As shown in Tab.~\ref{tab:segmethods}, the F-measures of these three methods are improved by 0.9\%, 1.3\%, and 0.9\% on CTW1500 respectively when the basic detectors are equipped with CBN. Similar conclusions can also be obtained on Total-Text, the F-measures of the basic detectors can be improved by 0.8\%, 1.0\%, and 0.8\% respectively. This indicates the generalization of our proposed CBN. In addition, the proposed CBN only increases a few parameters (about 0.01M) on the baseline model, and has little impact on speed, which all demonstrates CBN is a lightweight plugin that can be merged into multiple segmentation-based methods.

To further verify the generalization of the proposed CBN on different backbones, we also conduct ablation studies on CTW1500 and Total-Text datasets. We use PAN as baseline model and change the backbone and model structure with or without CBN, and train them from scratch with the default parameter settings. As shown in Tab.~\ref{tab:cbn_backbone}, the F-measure of these backbones with CBN can bring 0.6\%, 0.7\%, and 1.3\% improvements on CTW1500 respectively. In addition, these backbones with CBN can achieve 1.0\%, 1.7\%, and 1.0\% performance gains in terms of F-measure on Total-Text respectively.

\subsubsection{Influence of the Backbone}
To better analyze the capability of the proposed CBN, we replace the lightweight backbone (ResNet18) with a heavier backbone (ResNet50 and ResNet50-DCN), and test on three different datasets (CTW1500, Total-Text, and ICDAR-MLT19). As shown in Tab.~\ref{tab:backbone}, compared to the model with a lightweight backbone, the model with a heavier backbone can make about 0.6\% (CTW1500) and 1.4\% (Total-Text) improvements on precision. In addition, on the large-scale test set (ICDAR-MLT19), both ResNet50 and ResNet50-DCN can bring over 1.9\% and 3.0\% improvements on F-measure. This indicates that the proposed detector with a heavier backbone can achieve better performance, especially on large-scale datasets.

\begin{table}[t!]
	\centering
	\caption{Quantitative detection results on CTW1500. ``P", ``R", and ``F" represent precision, recall, and F-measure respectively.}
	\resizebox{\linewidth}{!}{
		\begin{tabular}{c|c|c|c|c}
			\hline
			\textbf{Method} & \textbf{Backbone} & \textbf{P(\%) }& \textbf{R \%)}&\textbf{F(\%)} \\
			\hline
			\hline
			TextSnake~\cite{long2018textsnake}  &VGG16    & 67.9 &85.3 &75.6  \\
			MSR~\cite{xue2019msr}  &Res50     & 85.0 & 78.3 &81.5  \\
			SegLink++~\cite{tang2019seglink++} &VGG16    & 82.8 & 79.8 &81.3  \\
			TextField~\cite{xu2019textfield}  & VGG16  & 83.0 & 79.8 & 81.4  \\
			TextMountain~\cite{zhu2021textmountain} &Res50 &82.9 &83.4 &83.2 \\ 
			PSENet~\cite{wang2019shape}  &Res50   & 84.8 & 79.7 &82.2  \\
			SAE~\cite{tian2019learning}  &Res50   &82.7 & 77.8  &80.1    \\
			LOMO~\cite{zhang2019look} &Res50   & 85.7 & 76.5 &80.8 \\
			CRAFT~\cite{baek2019character} &VGG16   & 86.0 & 81.1 &83.5    \\
			CRNet~\cite{2020CRNet}  & Res50   &87.0 &80.9&83.8  \\
			Feng \emph{et al.}~\cite{feng2021residual}  &VGG-16 & 87.3 &81.8 &84.5 \\
			PAN~\cite{wang2019efficient}&Res18  & 86.2 & 80.9 &83.5 \\
			PAN++~\cite{wang2021pan++} &Res18 &87.1 &81.1 &84.0 \\
			DRRGN~\cite{zhang2020deep}  &VGG16    & 85.9& 83.0&84.5   \\
			TextFuseNet~\cite{ye2020textfusenet}  & Res50  & 85.8 & 85.0 & 85.4  \\
			ContourNet~\cite{wang2020contournet}  &Res50   &83.7 &84.1  &83.9    \\
			ABCNet~\cite{liu2020abcnet}  &Res50  &84.4  &78.5 &81.4   \\
			ABCNetv2~\cite{liu2021abcnet}  &Res50  & 83.8 &85.6 &84.7  \\
			DBNet~\cite{liao2020real}  &Res50-DCN  & 86.9 & 80.2 &83.4   \\
			CT~\cite{sheng2021centripetaltext}  &Res18   &88.3 &79.9 &83.9  \\
			PCR~\cite{dai2021progressive}  &DLA34    & 87.2 &82.3 &84.7  \\
			FCENet~\cite{zhu2021fourier}  &Res50-DCN    &87.6  &83.4 &85.5   \\
			Raisi \emph{et al.}~\cite{raisi2022arbitrary} & Res50 & 89.2 & 85.3 & \textbf{87.2} \\
			TextBPN~\cite{zhang2021adaptive}  &Res50    &86.5 &83.6  &85.0  \\
			DBNet++~\cite{liao2022real} &Res50-DCN    &87.9 &82.8 &85.3  \\
			TextBPN++~\cite{zhang2022arbitrary}  &Res50    &87.3 &83.8  &85.5  \\
			Tang \emph{et al.}~\cite{tang2022few}  &Res50   &88.1 &82.4 &85.2    \\
			Long \emph{et al.}~\cite{long2022towards} &MaX-DeepLab    & 83.9 &\textbf{85.9} &84.9 \\
			\hline
			\textbf{Ours (Synth)}  &Res18   & 89.0 &81.9	&85.3   \\
			\textbf{Ours (MLT)}  &Res18   & \textbf{89.3} &82.9	&86.0   \\
			\hline
		\end{tabular}
	}
	\label{tab:ctw1500}
\end{table}

\begin{table}[t!]
	\centering
	\caption{Quantitative detection results on Total-Text. ``P", ``R", and ``F" represent precision, recall, and F-measure respectively.}
	\resizebox{\linewidth}{!}{
		\begin{tabular}{c|c|c|c|c}
			\hline
			\textbf{Method}  & \textbf{Backbone} & \textbf{P(\%) }& \textbf{R(\%)}&\textbf{F(\%)} \\
			\hline
			\hline
			TextSnake~\cite{long2018textsnake}&VGG16    & 82.7 & 74.5 & 78.4  \\
			MSR~\cite{xue2019msr}  &Res50      & 83.8 & 74.8 & 79.0  \\
			SegLink++~\cite{tang2019seglink++} &VGG16     & 82.1 & 80.9 & 81.5  \\
			TextField~\cite{xu2019textfield}  & VGG16  & 81.2 & 79.9 & 80.6   \\
			PSENet~\cite{wang2019shape}  &Res50    & 84.0 & 78.0 & 80.9   \\
			SPCNet~\cite{xie2019scene}  &Res50   & 83.0 & 82.8 & 82.9  \\
			Boundary~\cite{wang2020all}  &Res50    & 85.2 & 83.5 & 84.3 \\
			LOMO~\cite{zhang2019look}  &Res50    & 87.6 & 79.3 & 83.3  \\
			CRAFT~\cite{baek2019character}  &VGG16      & 87.6 & 79.9 & 83.6 \\
			CRNet~\cite{2020CRNet} & Res50   &85.8 &82.5&84.1  \\
			Feng \emph{et al.}~\cite{feng2021residual} &VGG-16 & 87.1 &80.3 &83.5 \\
			PAN~\cite{wang2019efficient}  &Res18  & 88.5 & 81.7& 85.0 \\
			PAN++~\cite{wang2021pan++} &Res18 &89.9 &81.0 &85.3 \\
			DRRGN~\cite{zhang2020deep} &VGG16     & 86.5  &84.9& 85.7  \\
			TextFuseNet~\cite{ye2020textfusenet}&Res50    &87.5 &83.2 &85.3     \\
			ContourNet~\cite{wang2020contournet}  &Res50     &86.9 &83.9  &85.4    \\
			ABCNet~\cite{liu2020abcnet}  & Res50  & 87.9 & 81.3 & 84.5  \\
			ABCNetv2~\cite{liu2021abcnet}  &Res50   &84.1 &90.2 &87.0   \\
			DBNet~\cite{liao2020real}  &Res50-DCN   & 87.1 & 82.5 & 84.7    \\
			CT~\cite{sheng2021centripetaltext}  &Res18    & 90.5 &82.5 &86.3  \\
			PCR~\cite{dai2021progressive}  &DLA34     &88.5 &82.0 &85.2  \\
			FCENet~\cite{zhu2021fourier}  &Res50-DCN     &89.3 &82.5  &85.8    \\
			Raisi \emph{et al.}~\cite{raisi2022arbitrary} & Res50 & 89.1 & 86.4 & 87.8 \\
			TextBPN~\cite{zhang2021adaptive}  &Res50     &90.7 & 85.2  &87.9  \\
			DBNet++~\cite{liao2022real}  &Res50-DCN     & 88.9 &83.2 &86.0 \\
			TextBPN++~\cite{zhang2022arbitrary}  &Res50   &\textbf{91.8} & 85.3  &\textbf{88.5}  \\
			Tang \emph{et al.}~\cite{tang2022few}  &Res50    &90.7 &85.7 &88.1   \\
			Long \emph{el al.}~\cite{long2022towards}  &MaX-DeepLab    & 85.5 &\textbf{90.5} &87.9  \\
			\hline 
			\textbf{Ours (Synth)}  &Res18   & 90.1	&82.5	&86.1  \\
			\textbf{Ours (MLT)}  &Res18   & 89.3 &85.2 &87.2  \\
			\hline
			
		\end{tabular}
	}
	\label{tab:totaltext}
\end{table}

\begin{table}[h!]
	\centering
	\caption{Quantitative detection results on MSRA-TD500. ``P", ``R", and ``F" represent precision, recall, and F-measure respectively.}
	\resizebox{\linewidth}{!}{
		\begin{tabular}{c|c|c|c|c}
			\hline
			\textbf{Method}  & \textbf{Backbone} & \textbf{P(\%) }& \textbf{R(\%)}&\textbf{F(\%)} \\
			\hline
			\hline
			SegLink~\cite{shi2017detecting} &VGG16  &86.0 &70.0 &77.0 \\
			TextSnake~\cite{long2018textsnake}  &VGG16     & 83.2 & 73.9 & 78.3 \\
			TextField~\cite{xu2019textfield}  & VGG16  & 87.4 & 75.9 & 81.3 \\
			SAE~\cite{tian2019learning}  &Res50   &84.2 & 81.7  &82.9  \\
			CRAFT~\cite{baek2019character}  &VGG16   & 88.2 & 75.9 & 82.9 \\
			MSR~\cite{xue2019msr}   &Res50 & 87.4 & 76.7 & 81.7  \\
			ATTR~\cite{2019Arbitrary} &SE-VGG16  &85.2 &82.1 & 83.6  \\
			MaskTextSpotterV3~\cite{liao2020mask} &Res50 &90.7 &77.5 &83.5 \\
			PAN~\cite{wang2019efficient}   &Res18  &85.7 & 83.4 & 84.5 \\
			PAN++~\cite{wang2021pan++} &Res18 &85.3 &84.0 &84.7 \\
			MCN~\cite{liu2020bottom} &Res50 & 89.1 &80.7 & 85.2 \\
			DRRGN~\cite{zhang2020deep} &VGG16    &88.1 & 82.3 & 85.1  \\
			ABCNetv2~\cite{liu2021abcnet}  &Res50  &81.3 &89.4 &85.2  \\
			DBNet~\cite{liao2020real}  &Res50-DCN   &91.5 & 79.2 & 84.9  \\
			MOST~\cite{he2021most} &Res50   & 90.4 & 82.7 & 86.4  \\
			CT~\cite{sheng2021centripetaltext} &Res18    &90.0 &82.5 &86.1  \\
			PCR~\cite{dai2021progressive} &DLA34    & 90.8 & 83.5 & 87.0  \\
			TextBPN~\cite{zhang2021adaptive} &Res50     &86.6 &84.5  &85.6  \\
			DBNet++~\cite{liao2022real} &Res50-DCN    &91.5 &83.3 &87.2 \\
			TextBPN++~\cite{zhang2022arbitrary} &Res50     &89.2 &85.4  &87.3  \\
			Raisi \emph{et al.}~\cite{raisi2022arbitrary} & Res50 & 88.1 & 85.0 &86.5 \\
			Tang \emph{et al.}~\cite{tang2022few}  &Res50     &91.4 &84.7 &87.9   \\
			Long \emph{et al.}~\cite{long2022towards} &MaX-DeepLab    &86.6 &\textbf{86.8} &86.7 \\
			\hline
			\textbf{Ours (Synth)}   &Res18   &91.1	&84.8	&87.8  \\
			\textbf{Ours (MLT)}   &Res18   &\textbf{93.2}	&86.4	&\textbf{89.7}  \\
			\hline
			
		\end{tabular}
	}
	\label{tab:msra}
\end{table}

\begin{table}[t!]
	\centering
	\caption{Quantitative detection results on COCO-Text test set. ``P", ``R", and ``F" represent precision, recall, and F-measure respectively. ``*" indicates that the results are on the competition website.}
	\resizebox{\linewidth}{!}{
		\begin{tabular}{c|c|c|c|c}
			\hline
			\textbf{Method} & \textbf{Backbone} & \textbf{P(\%) }& \textbf{R \%)}&\textbf{F(\%)} \\
			\hline
			\hline
		
			TextBoxes++~\cite{liao2018textboxes++} &VGG-16 &60.9 &56.7  &58.7  \\
			Lyu \emph{et al.}~\cite{lyu2018multi} & VGG-16 &\textbf{72.5} &52.9 &61.1 \\
			Mask TextSpotter~\cite{lyu2018mask} &Res50 & 66.8 &58.3 &62.3 \\
			UHT~\cite{wang2020method} &Res50 & 60.8 &49.0  &54.2 \\
			Wang \emph{et al.}~\cite{wang2020all} &Res50 &59.0 &67.7 &63.0 \\
			Feng \emph{et al.}~\cite{feng2021residual} &Res50 &66.8 &59.4 &62.9 \\
			FC$^2$RN~\cite{qin2021fc} & Res50 & 68.5 & 58.2 &63.0 \\
			Text\_Detection\_DL* &- & 60.9 & 61.8 & 61.4 \\
			Test-Msk\_v2 * & - & 59.6 & \textbf{68.7} & 63.8 \\
			
			\hline
			\textbf{Ours (MLT)} &Res50 &62.1 & 66.8 & \textbf{64.4} \\
			\hline
		\end{tabular}
	}
	\label{tab:cocotext}
\end{table}

\begin{table}[t!]
	\centering
	\caption{Quantitative detection results on ICDAR 2015. ``P", ``R" and ``F" represent precision, recall, and F-measure respectively.}
	\resizebox{\linewidth}{!}{
		\begin{tabular}{c|c|c|c|c}
			\hline
			\textbf{Method} & \textbf{Backbone} & \textbf{P(\%) }& \textbf{R \%)}&\textbf{F(\%)} \\
			\hline
			\hline
			TextSnake~\cite{long2018textsnake}  &VGG16    &84.9 &80.4 &82.6  \\
			Mask TextSpotter~\cite{lyu2018mask}  &Res50  &91.6 &81.0 &86.0 \\
			MSR~\cite{xue2019msr}  &Res50     & 86.6 &78.4 &82.3   \\
			SegLink++~\cite{tang2019seglink++} &VGG16   &83.7  & 80.3  &82.0  \\
			TextField~\cite{xu2019textfield}  & VGG16  & 84.3 &83.9 &84.1 \\
			
			TextMountain~\cite{zhu2021textmountain} &Res50 &88.5 &84.2 &86.3  \\ 
			PSENet~\cite{wang2019shape}  &Res50   & 86.9 &84.5 &85.7  \\
			SAE~\cite{tian2019learning}  &Res50    &88.3 &85.0 &86.6   \\
			LOMO~\cite{zhang2019look} &Res50   & 91.3 &83.5 &87.2 \\
			
			CRAFT~\cite{baek2019character} &VGG16   & 89.8 &84.3 &86.9    \\

			CRNet~\cite{2020CRNet}  & Res50   &88.3 &84.5  &86.4  \\
			
			PAN~\cite{wang2019efficient}&Res18  & 84.0 &81.9 &82.9 \\
			
			PAN++~\cite{wang2021pan++} &Res50 &91.4 &83.9 &87.5 \\
			DRRGN~\cite{zhang2020deep}  &VGG16    & 88.5 &84.7 &86.6   \\
			
			ContourNet~\cite{wang2020contournet}  &Res50   &87.6 &86.1  &86.9    \\
			
			DBNet~\cite{liao2020real}  &Res50-DCN  & \textbf{91.8} &83.2 &87.3   \\
			
			MOST~\cite{he2021most} &Res50 &89.1 &\textbf{87.3} &88.2 \\
			FCENet~\cite{zhu2021fourier}  &Res50    &90.1  &82.6 &86.2   \\
			Raisi \emph{et al.}~\cite{raisi2022arbitrary} & Res50 & 90.2 & 83.1 &86.5 \\
			DBNet++~\cite{liao2022real} &Res50-DCN    &90.9 &83.9 &87.3 \\
			Tang \emph{et al.}~\cite{tang2022few}  &Res50   &91.1 &86.7 &\textbf{88.8}  \\
			\hline
			\textbf{Ours (MLT)}  &Res18   & 90.3 &84.4	&87.3   \\
			\textbf{Ours (MLT)}  &Res50   & 89.0 &85.5	&87.2   \\
			\textbf{Ours (MLT)}  &Res50-DCN   &91.0 &85.4	&88.1   \\
			\hline
		\end{tabular}
	}
	\label{tab:icdar15}
\end{table}

\subsection{Comparisons with State-of-the-Art Methods}
To prove the superiority of our method, we make a comparison with other state-of-the-art methods on five public benchmarks. As PAN~\cite{wang2019efficient} has superior performance in the existing segmentation-based methods, we choose PAN-1s as the basic detector in this subsection. If not specified, the evaluation protocol for each dataset is the same as the basic detector.
	
\begin{figure*}[t!]
	\centering
	\begin{subfigure}[b]{\textwidth}
		\centering
		\includegraphics[width=.24\textwidth]{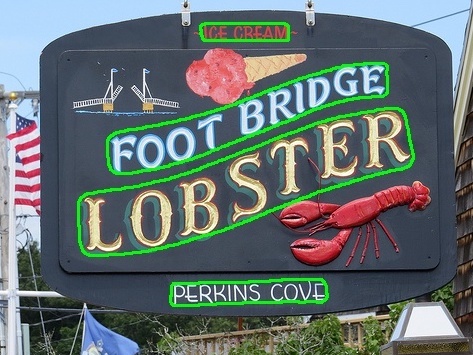}
		\hspace{.2pt}
		\includegraphics[width=.24\textwidth]{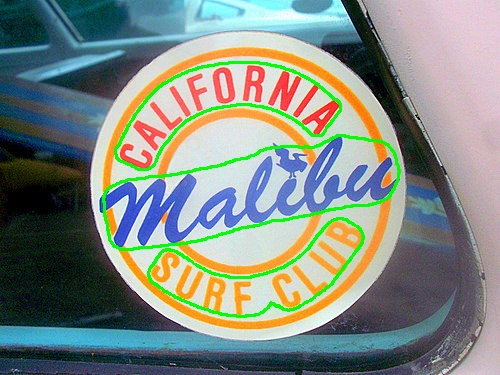}
		\hspace{.2pt}
		\includegraphics[width=.24\textwidth]{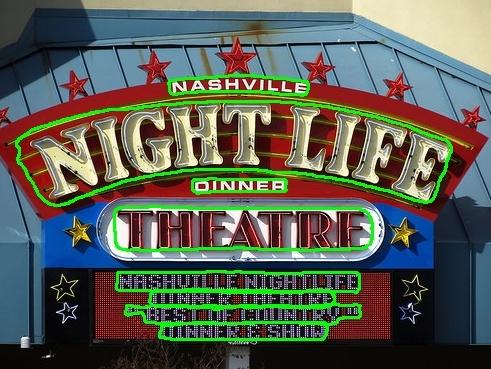}
		\hspace{.2pt}
		\includegraphics[width=.24\textwidth]{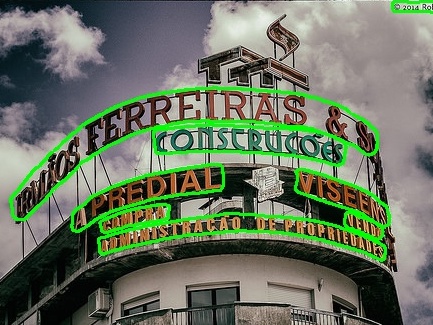}
		\vspace{3.5pt}
		\label{subfig:ctwshow}
	\end{subfigure}
	\begin{subfigure}[b]{\textwidth}
		\centering
		\includegraphics[width=.24\textwidth]{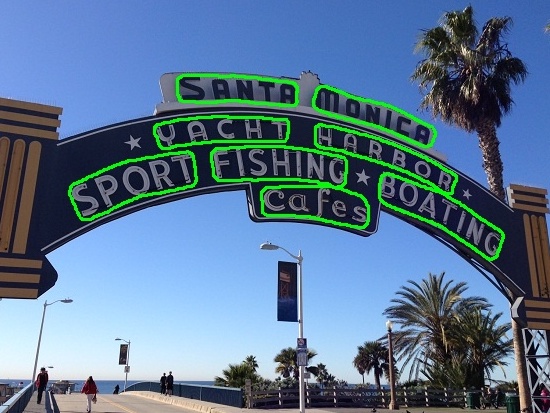}
		\hspace{.2pt}
		\includegraphics[width=.24\textwidth]{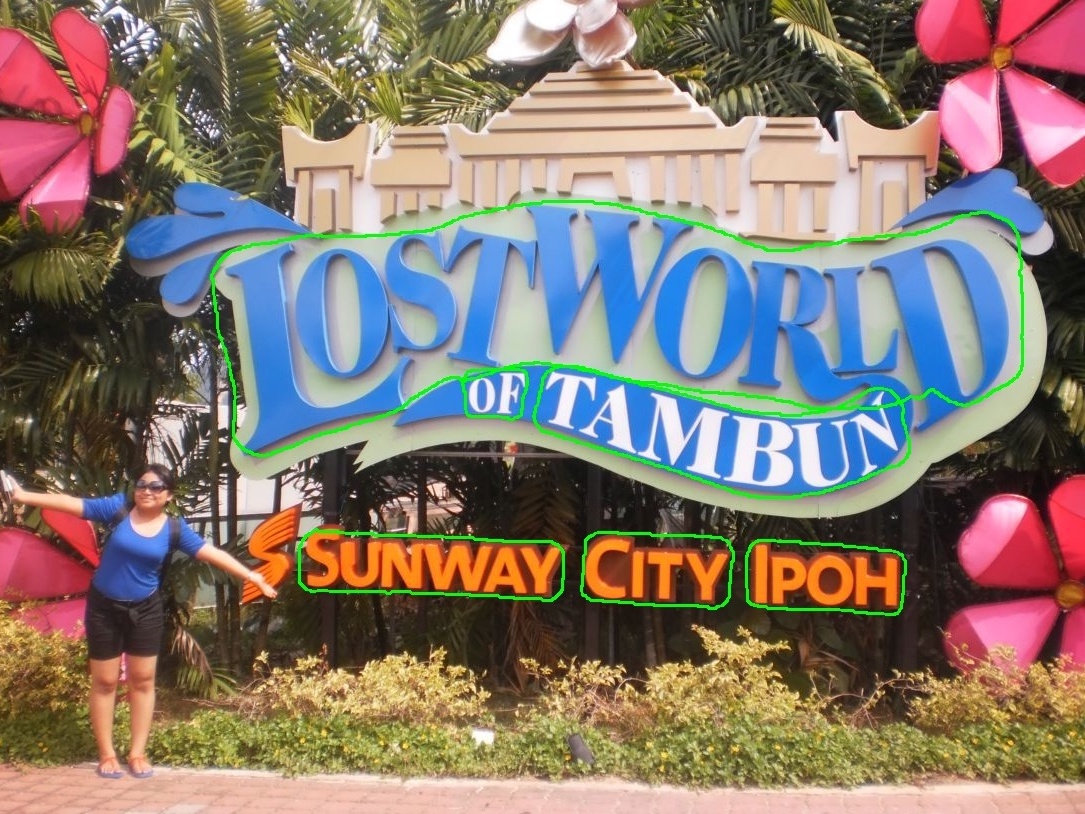}
		\hspace{.2pt}
		\includegraphics[width=.24\textwidth]{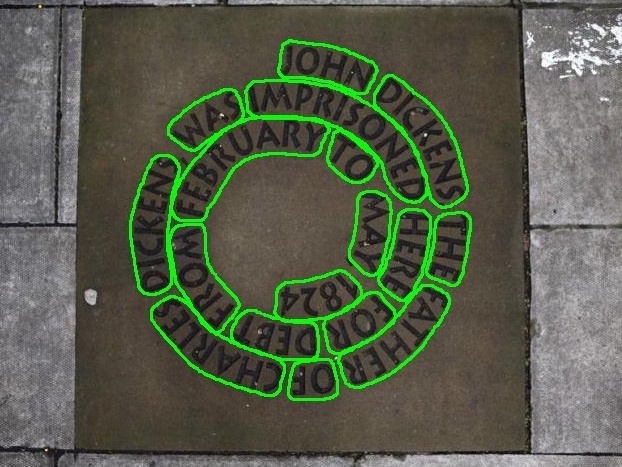}
		\hspace{.2pt}
		\includegraphics[width=.24\textwidth]{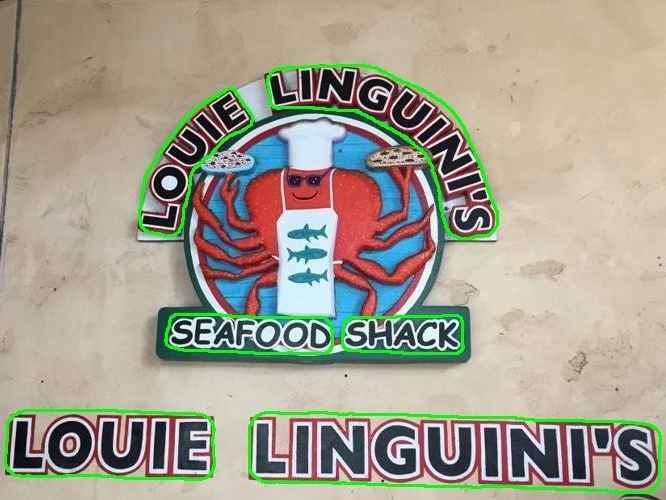}
		\label{subfig:ttshow}
	\end{subfigure}
\caption{Visualization results of our CBN on curve datasets. The first and second rows are CTW1500 and Total-Text datasets respectively.}
\label{fig:visualshowa}
\end{figure*}

\begin{figure*}[t!]
	\centering
	\begin{subfigure}[b]{\textwidth}
		\centering
		\includegraphics[width=.24\textwidth]{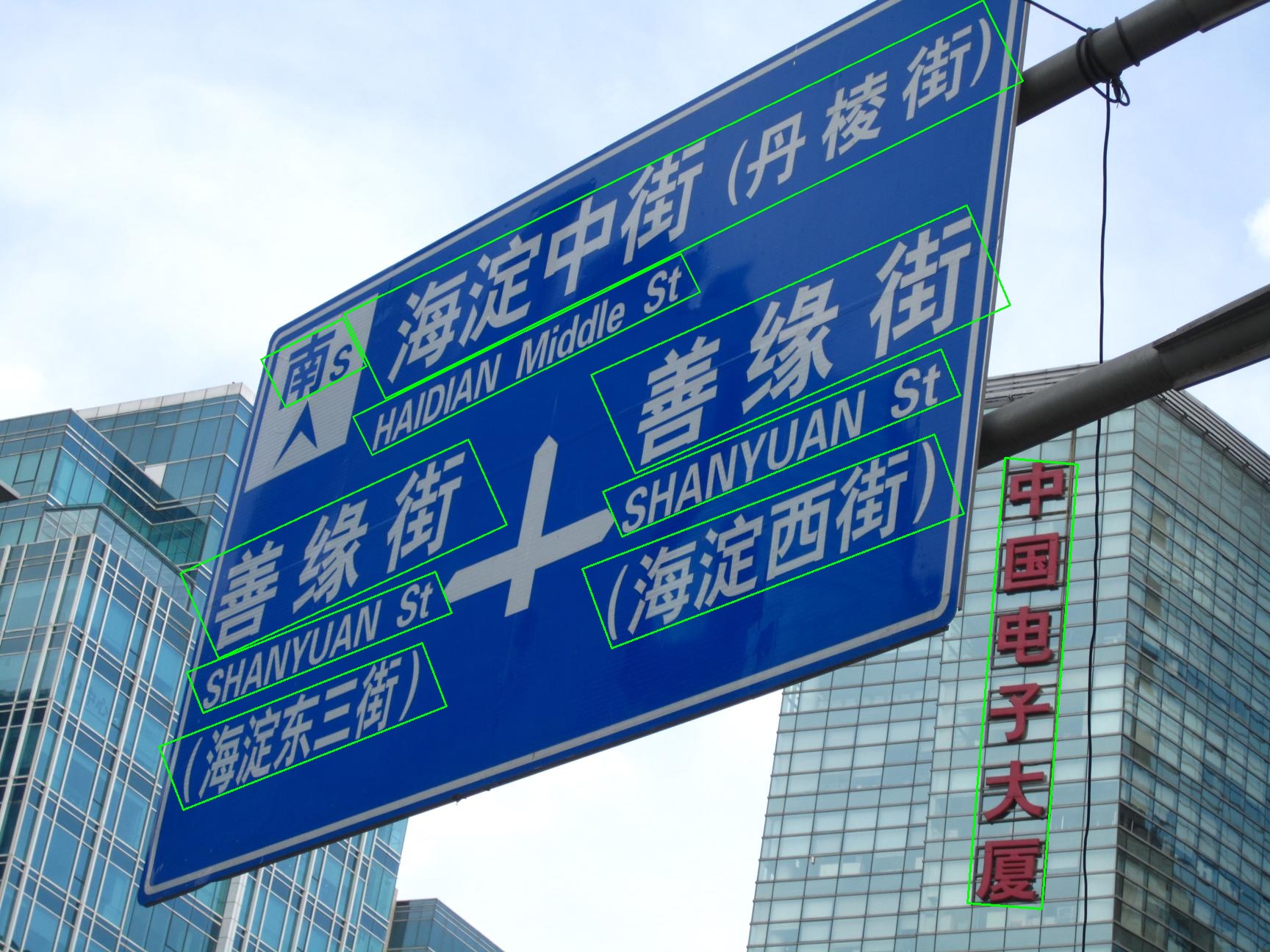}
		\vspace{2.pt}
		\includegraphics[width=.24\textwidth]{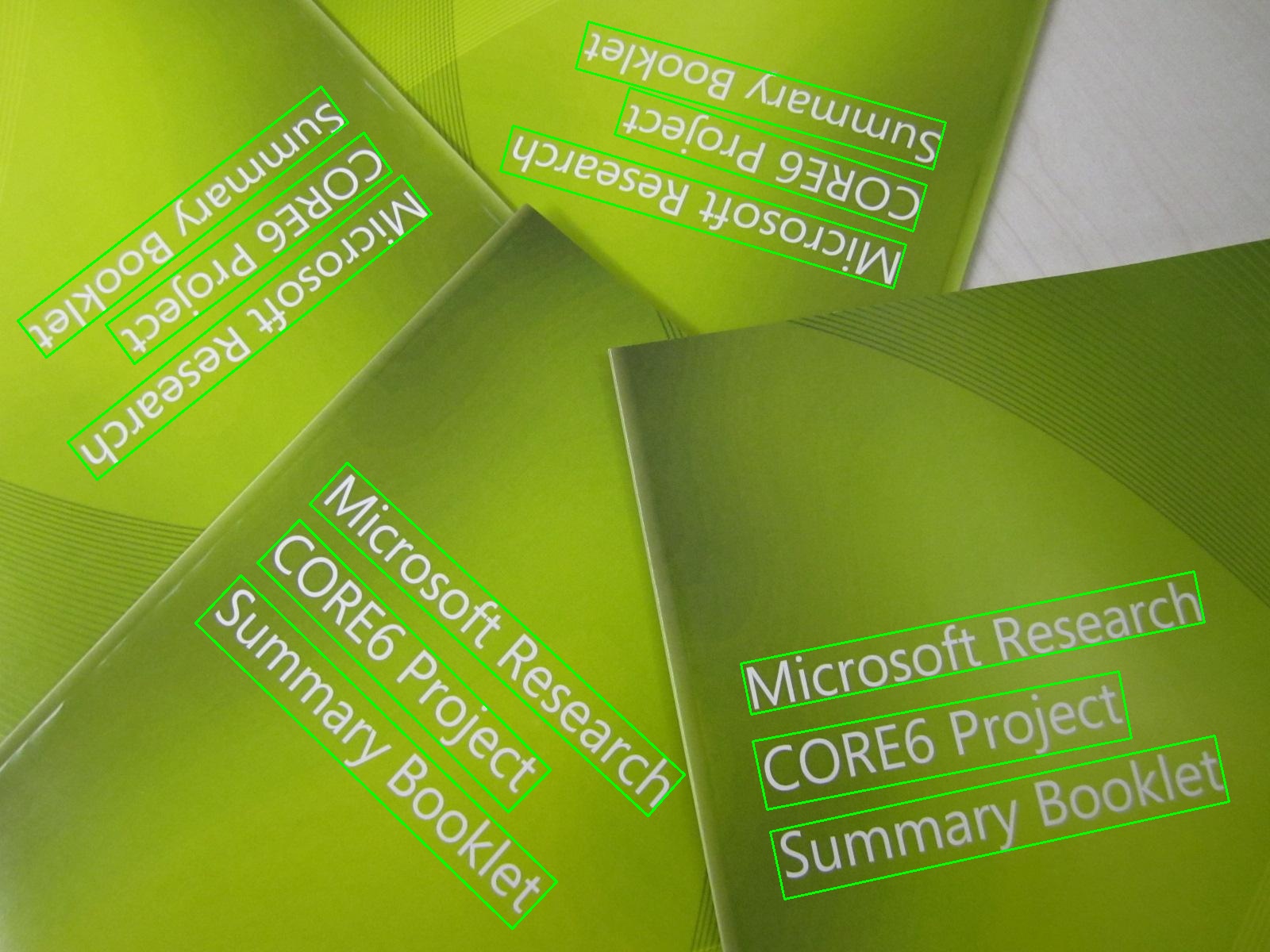}
		\vspace{2.pt}
		\includegraphics[width=.24\textwidth]{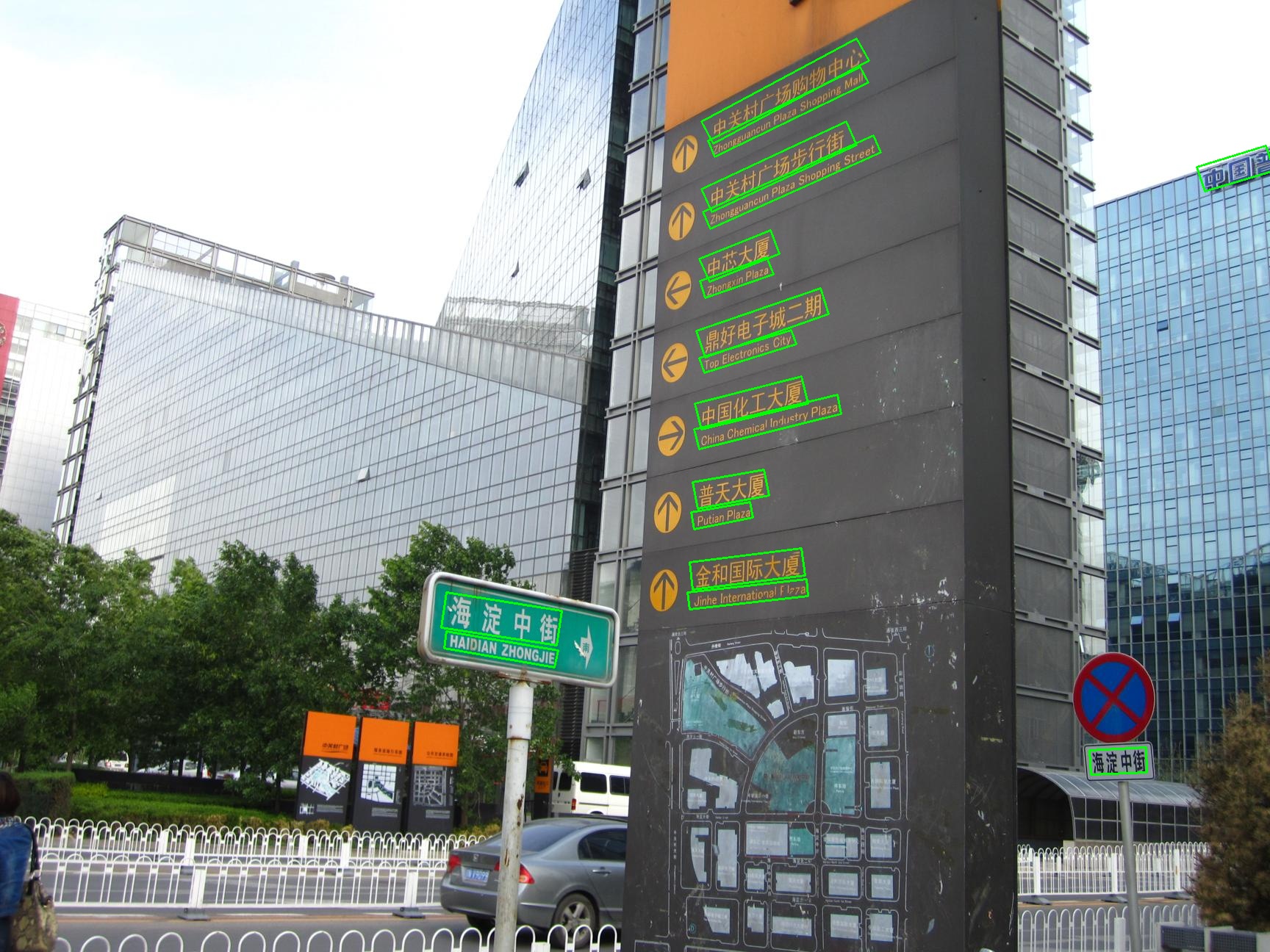}
		\vspace{2.pt}
		\includegraphics[width=.24\textwidth]{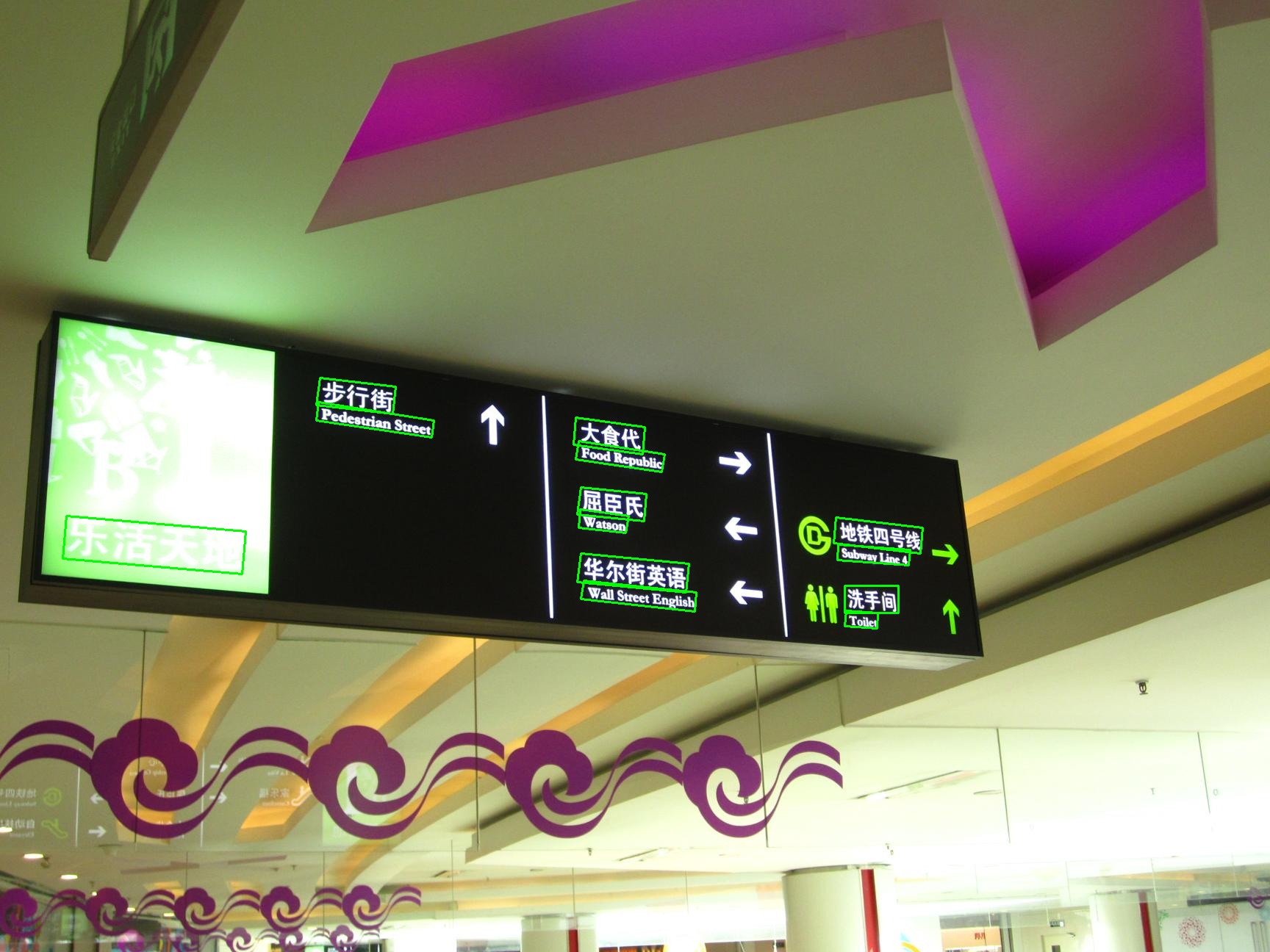}
		\vspace{.6pt}
		\label{subfig:msrashow}
	\end{subfigure}
	\begin{subfigure}[b]{\textwidth}
		\centering
		\includegraphics[width=.24\textwidth]{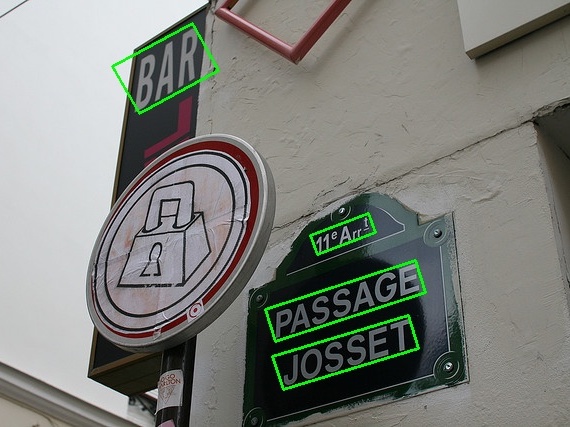}
		\vspace{2.pt}
		\includegraphics[width=.24\textwidth]{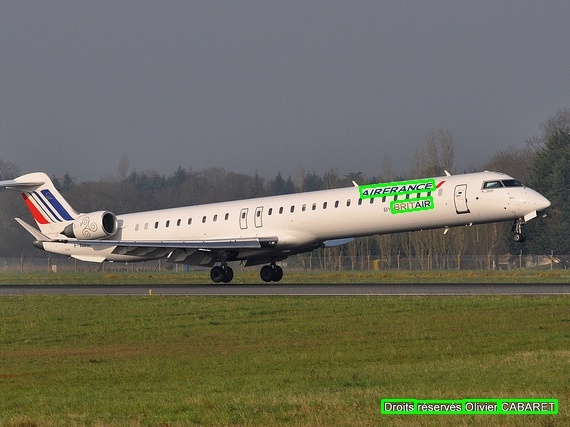}
		\vspace{2.pt}
		\includegraphics[width=.24\textwidth]{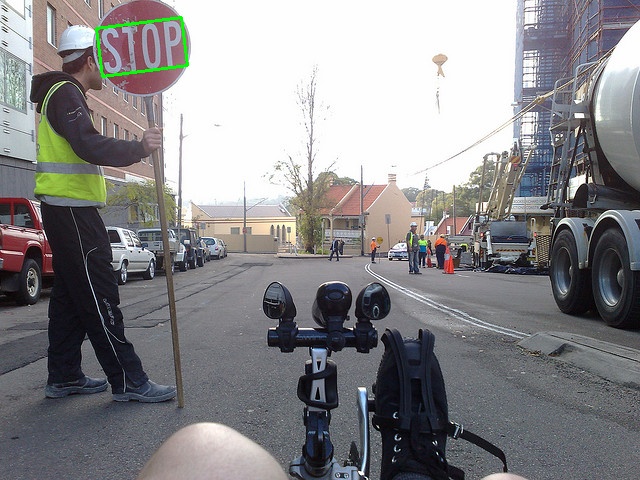}
		\vspace{2.pt}
		\includegraphics[width=.24\textwidth]{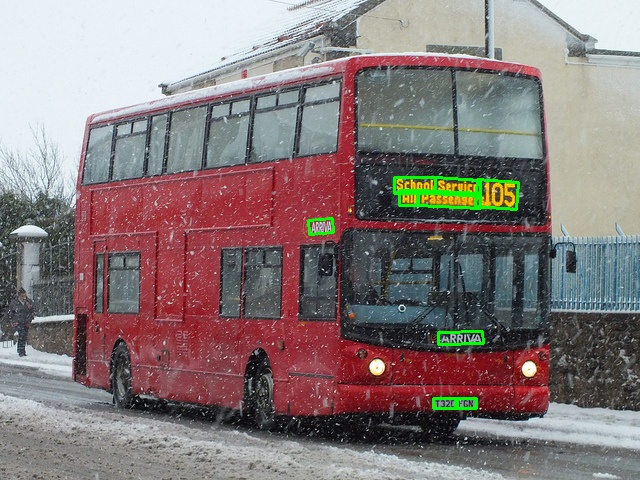}
		\vspace{.6pt}
		\label{subfig:cocoshow}
	\end{subfigure}
	\begin{subfigure}[b]{\textwidth}
		\centering
		\includegraphics[width=.24\textwidth]{img_10.jpg}
		\vspace{2.pt}
		\includegraphics[width=.24\textwidth]{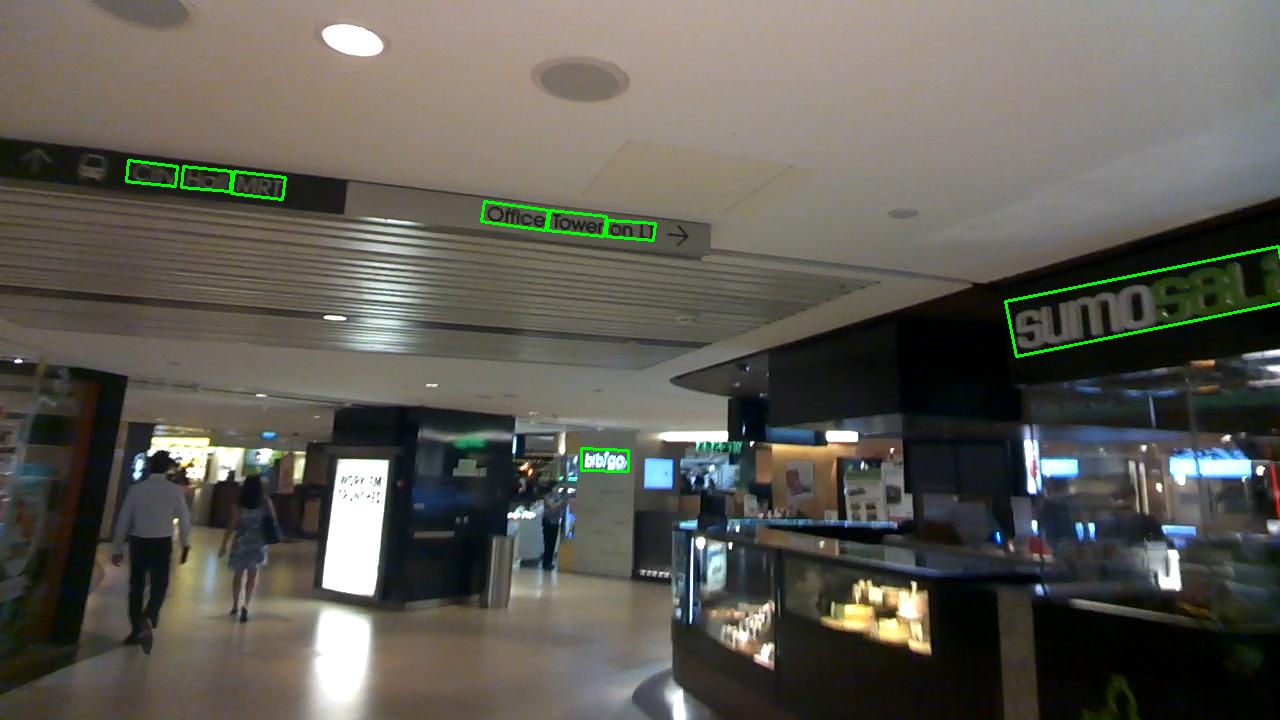}
		\vspace{2.pt}
		\includegraphics[width=.24\textwidth]{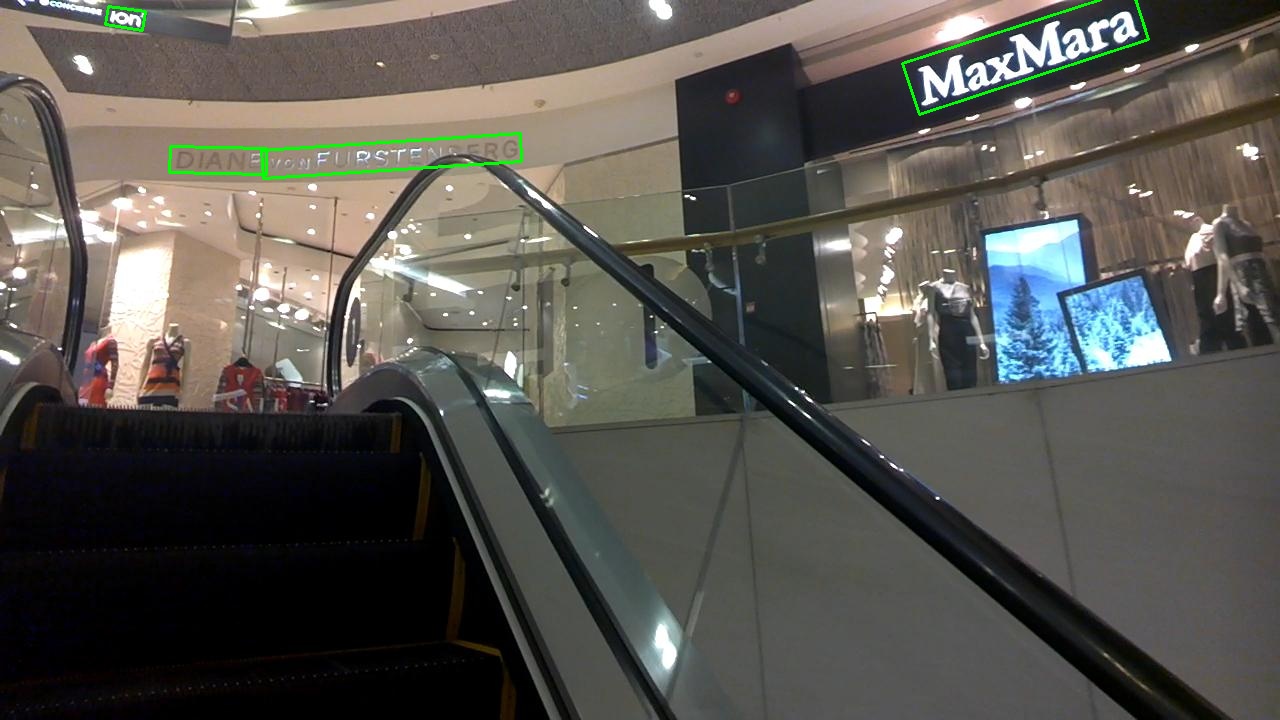}
		\vspace{2.pt}
		\includegraphics[width=.24\textwidth]{img_333.jpg}
		\vspace{.6pt}
		\label{subfig:ic15show}
	\end{subfigure}
	\begin{subfigure}[b]{\textwidth}
		\centering
		\includegraphics[width=.24\textwidth]{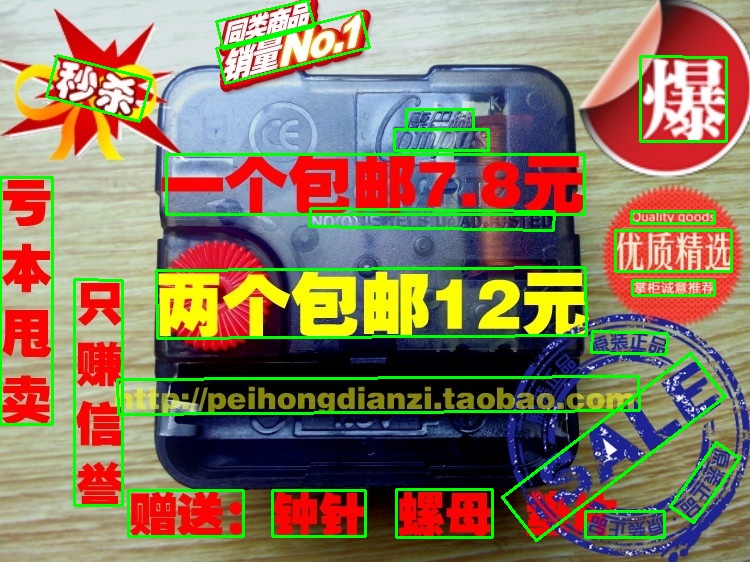}
		\vspace{2.pt}
		\includegraphics[width=.24\textwidth]{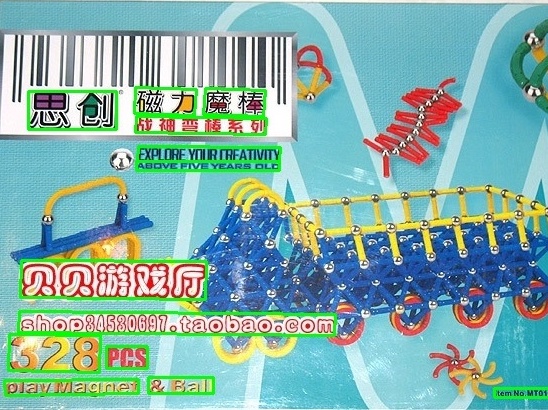}
		\vspace{2.pt}
		\includegraphics[width=.24\textwidth]{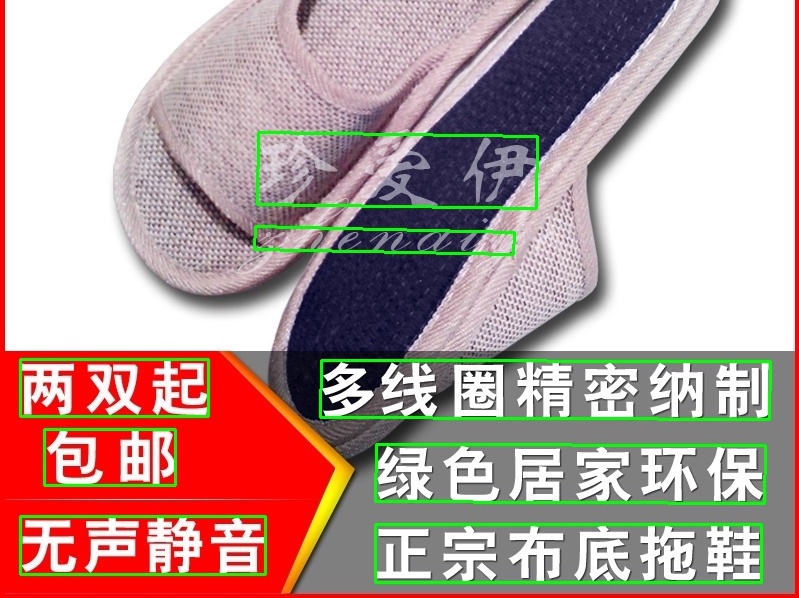}
		\vspace{2.pt}
		\includegraphics[width=.24\textwidth]{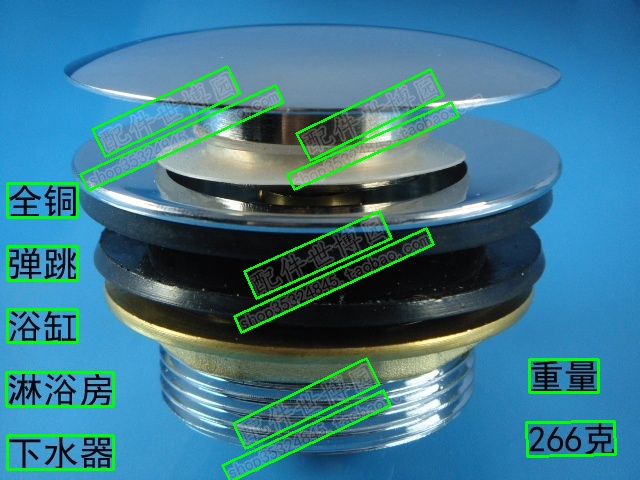}
		\vspace{.6pt}
		\label{subfig:mtwishow}
	\end{subfigure}
	\begin{subfigure}[b]{\textwidth}
		\centering
		\includegraphics[width=.24\textwidth]{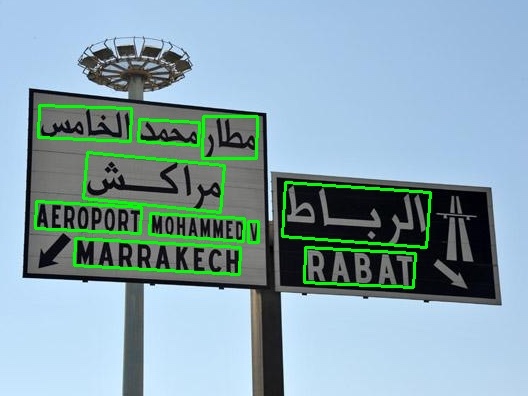}
		\vspace{2.pt}
		\includegraphics[width=.24\textwidth]{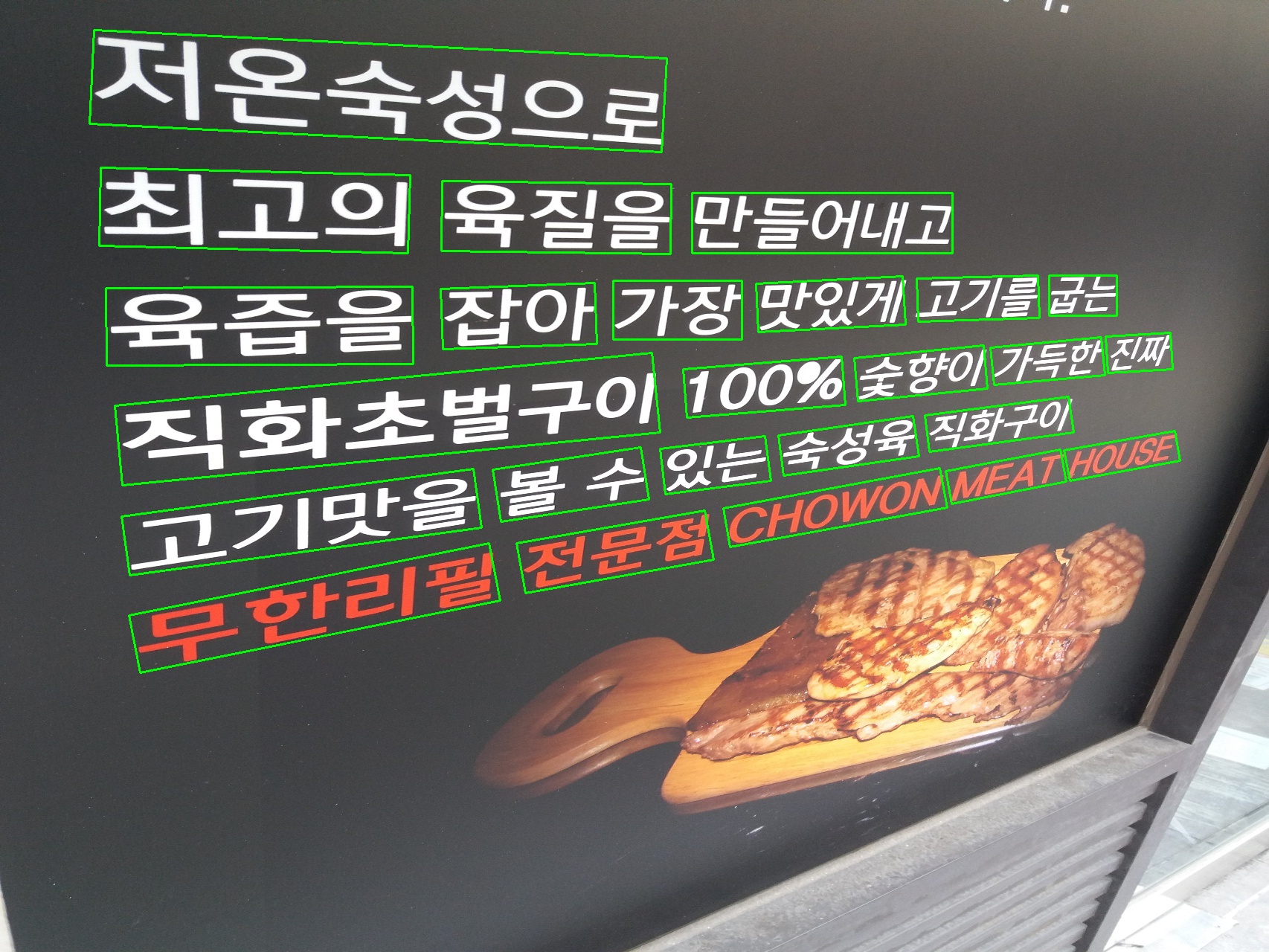}
		\vspace{2.pt}
		\includegraphics[width=.24\textwidth]{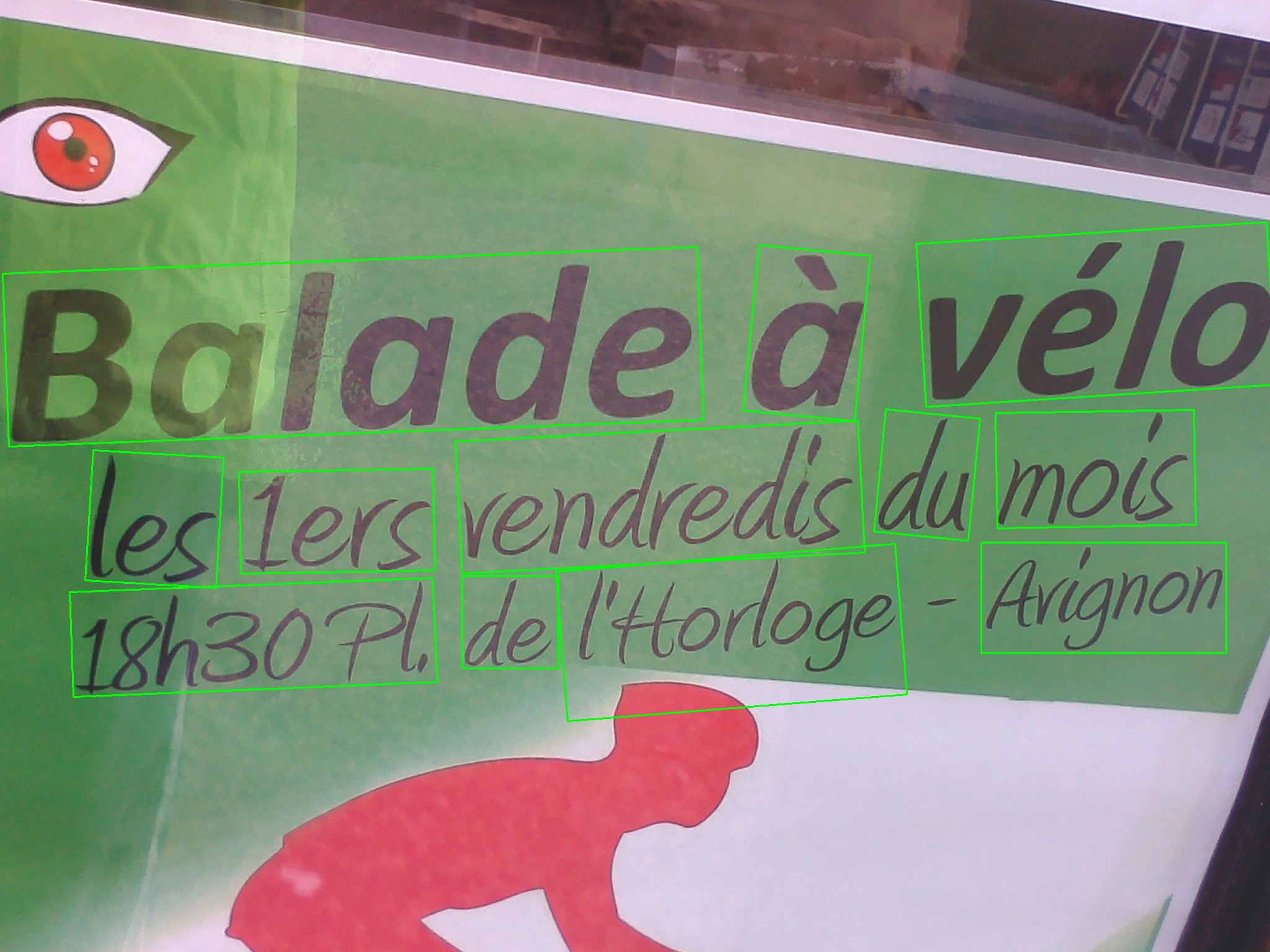}
		\vspace{2.pt}
		\includegraphics[width=.24\textwidth]{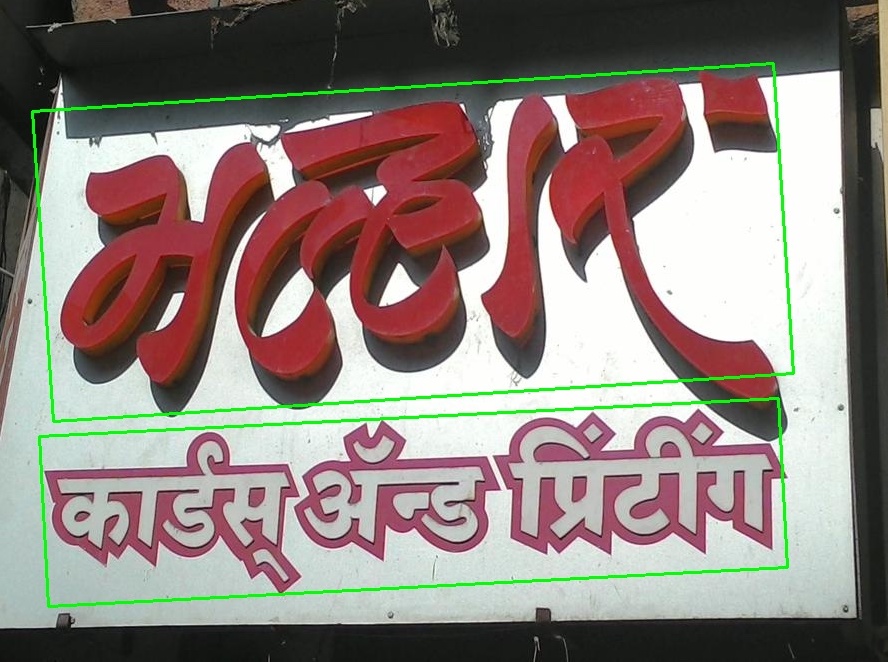}
		\label{subfig:mlt19show}
	\end{subfigure}
	\caption{Visualization results of our CBN on multi-oriented and multi-lingual datasets. The first to fifth rows are MSRA-TD500, COCO-Text, ICDAR 2015, MTWI, and ICDAR-MLT19 datasets respectively.}
	\label{fig:visualshowb}
\end{figure*}

\subsubsection{Curve Text Detection}
To test the ability of curve text detection, we compare the performance of our method with other state-of-the-art methods on CTW1500 and Total-Text which contain many curve text instances. During testing, we keep the aspect ratio of the test images and resize the short side of images to 640. As shown in Tab.~\ref{tab:ctw1500} and~\ref{tab:totaltext}, our results are lower than Raisi \emph{et al.}'s method~\cite{raisi2022arbitrary} on CTW1500, but our method can keep high speed and achieve a competitive F-measure of and 86.0\% on CTW1500. Although our result is lower than other SOTA methods on Total-Text, our method has a low computational cost and a competitive inference speed due to the lightweight backbone (ResNet18). In summary, the performances on CTW1500 and Total-Text demonstrate the solid superiority of the proposed CBN to detect arbitrary-shaped text instances. We also visualize our detection results in Fig.~\ref{fig:visualshowa}, which shows that our method can handle arbitrary shape texts well.

\begin{figure*}[t!]
	\centering
	\begin{subfigure}[b]{.32\textwidth}
		\centering
		\includegraphics[width=\textwidth]{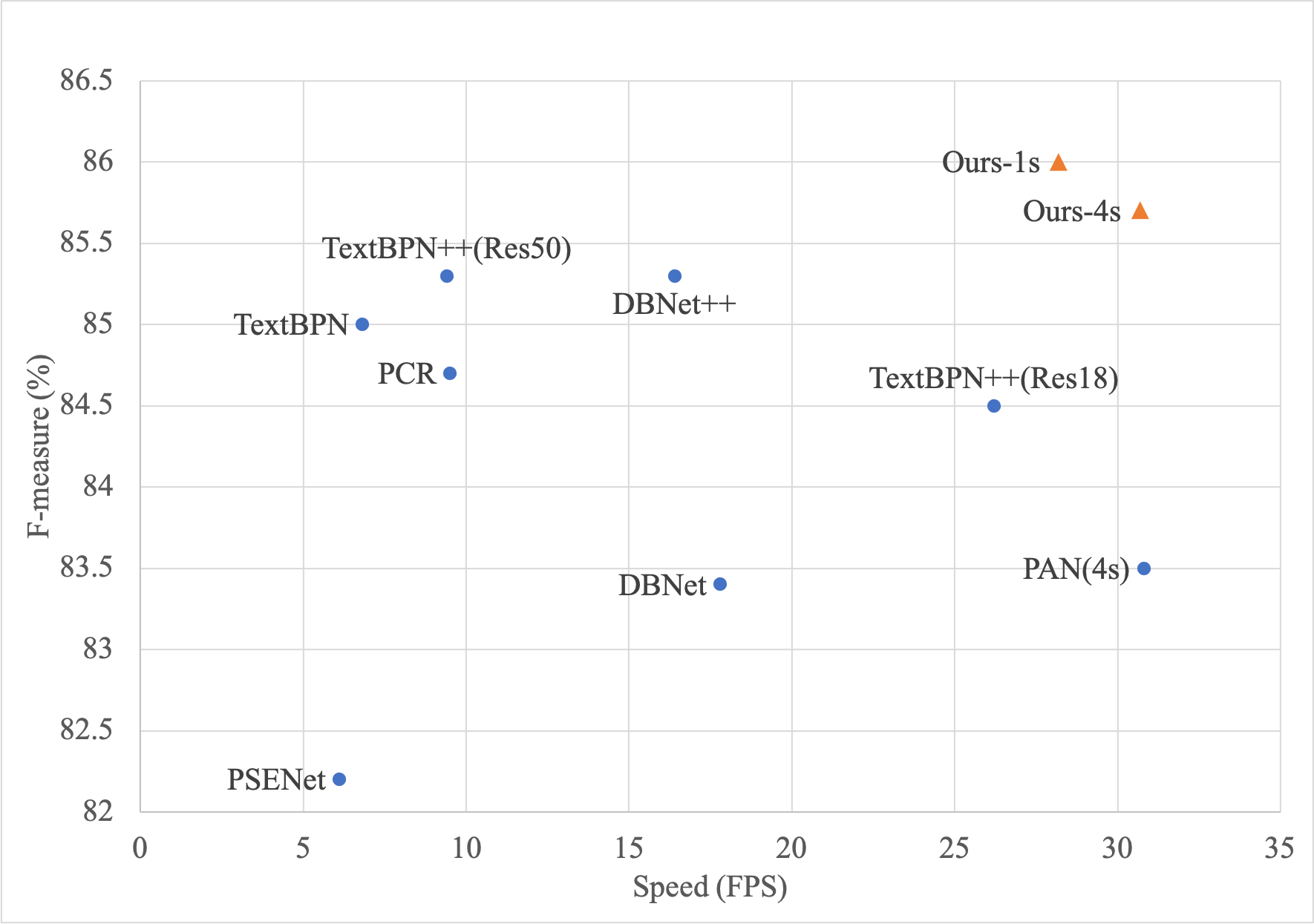}
		\caption{}
		\label{subfig:speed}
	\end{subfigure}
	\hspace{-1.5pt}	
	\begin{subfigure}[b]{.32\textwidth}
		\centering
		\includegraphics[width=\textwidth]{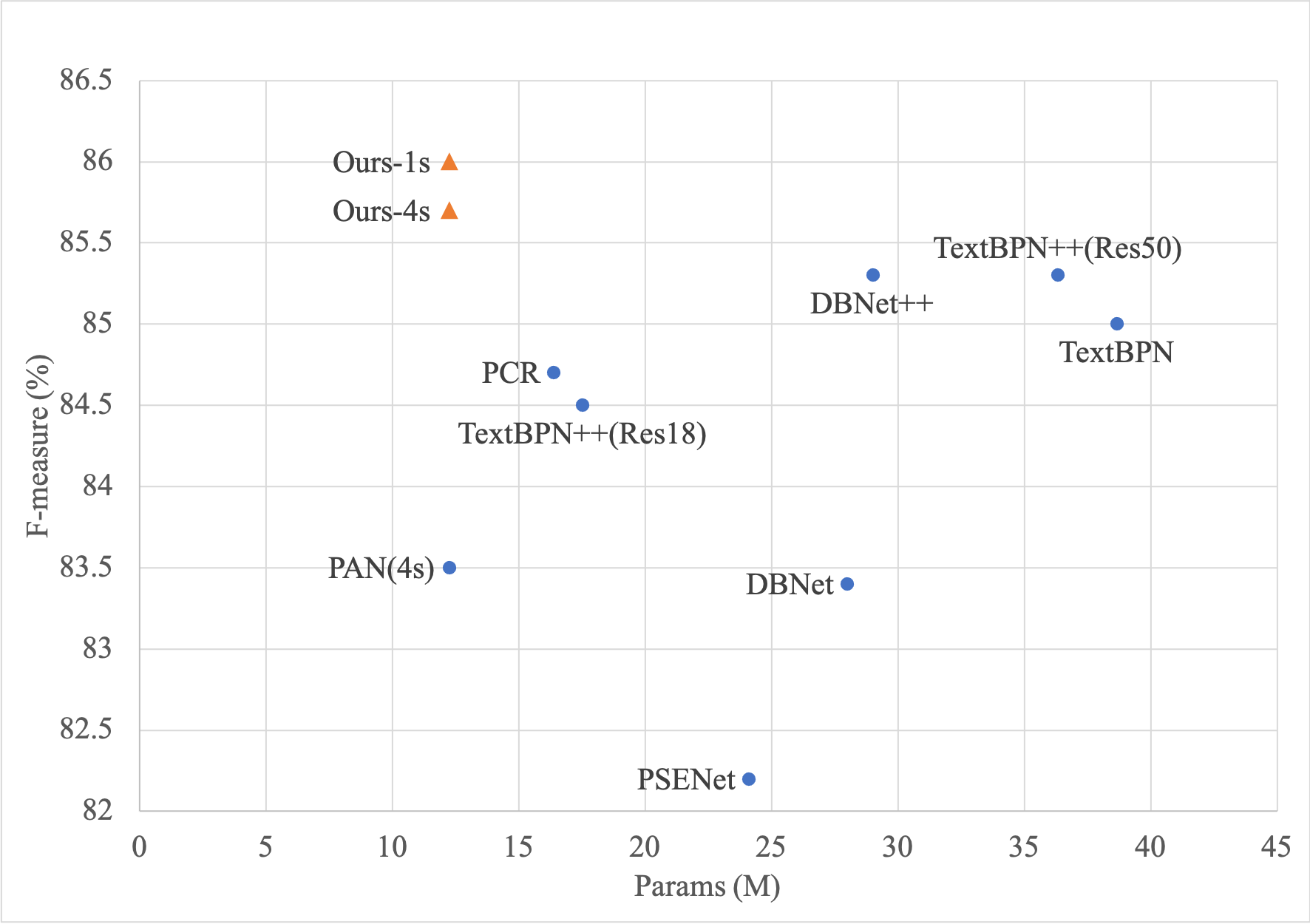}
		\caption{}
		\label{subfig:params}
	\end{subfigure}
	\hspace{-1.5pt}	
	\begin{subfigure}[b]{.32\textwidth}
		\centering
		\includegraphics[width=\textwidth]{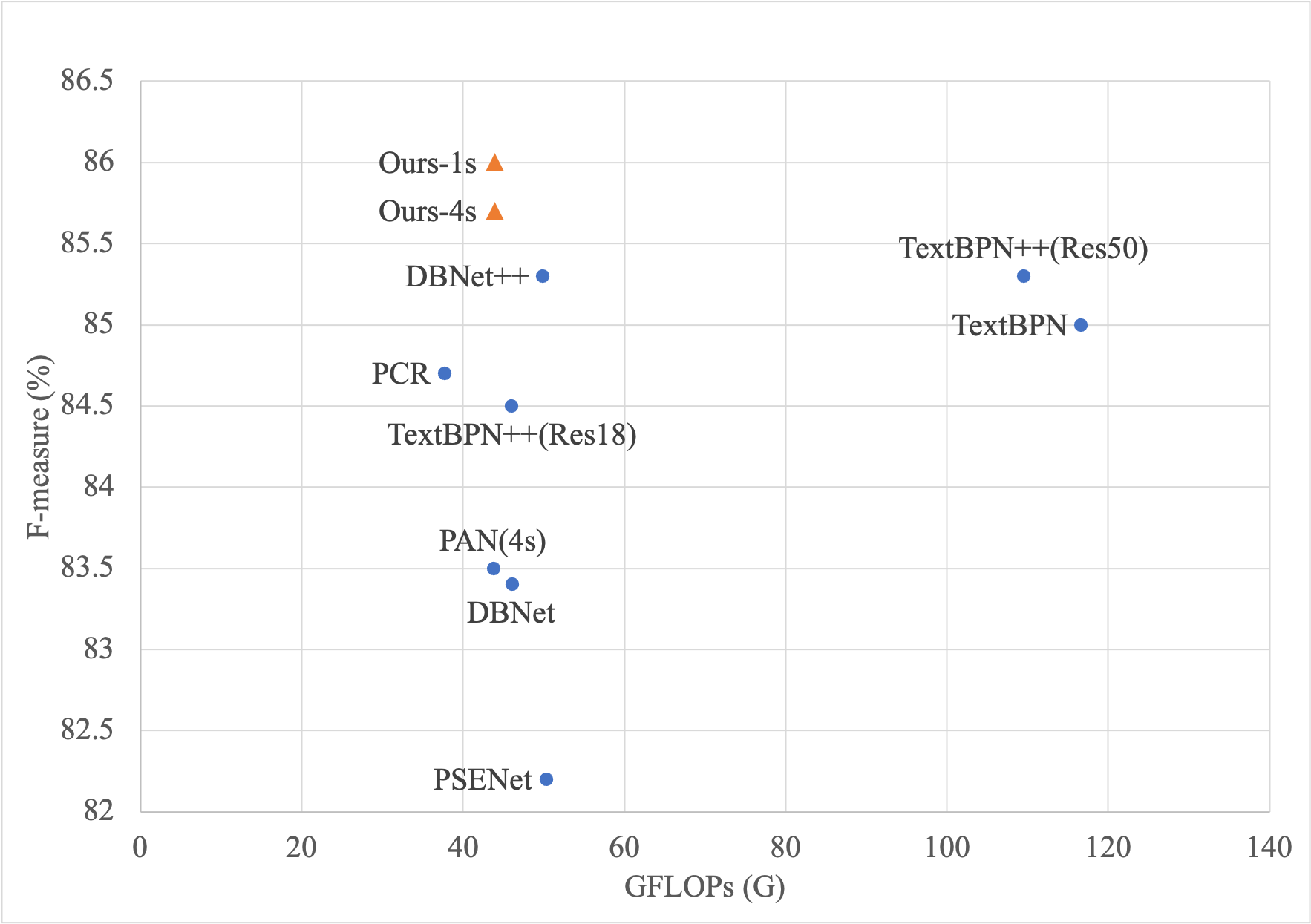}
		\caption{}
		\label{subfig:gflops}
	\end{subfigure}
	\caption{The performance and three common cost metrics (FPS, number of parameters, and GFLOPs) on CTW1500 dataset.}
	\label{fig:ctwperformance}
\end{figure*}

\begin{figure*}[t!]
	\centering
	\begin{subfigure}{.235\linewidth}
		\centering
		\includegraphics[width=\linewidth]{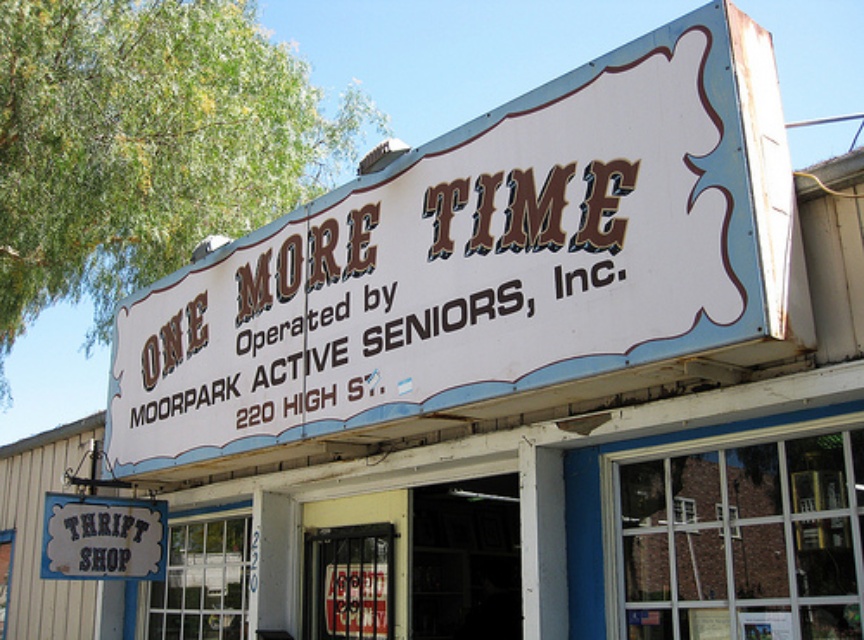}
		\caption{Input image}
		\label{subfig:input}
	\end{subfigure}
	\hspace{-1pt}
	\begin{subfigure}{.235\linewidth}
		\centering
		\includegraphics[width=\linewidth]{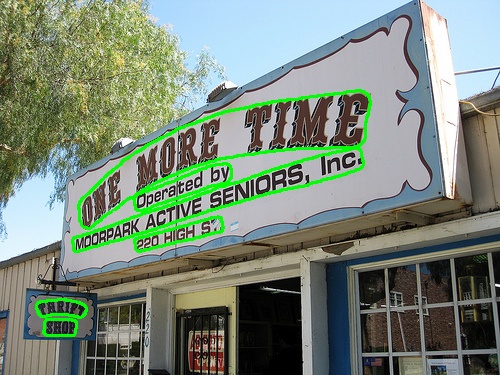}
		\caption{Detection result}
		\label{subfig:detection}
	\end{subfigure}
	\hspace{-1pt}
	\begin{subfigure}{.235\linewidth}
		\centering
		\includegraphics[width=\linewidth]{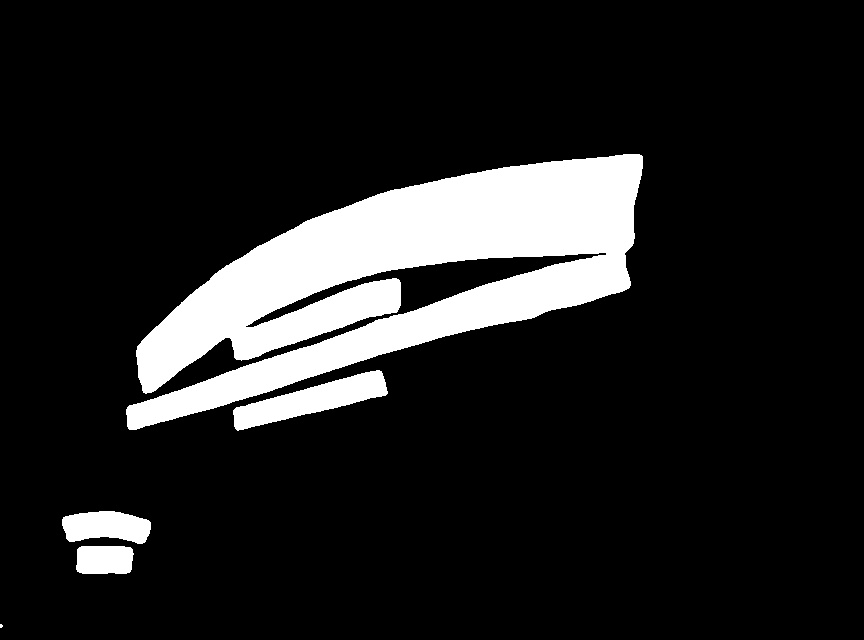}
		\caption{Text region map}
		\label{subfig:textregion}
	\end{subfigure}
	\hspace{-1pt}
	\begin{subfigure}{.235\linewidth}
		\centering
		\includegraphics[width=\linewidth]{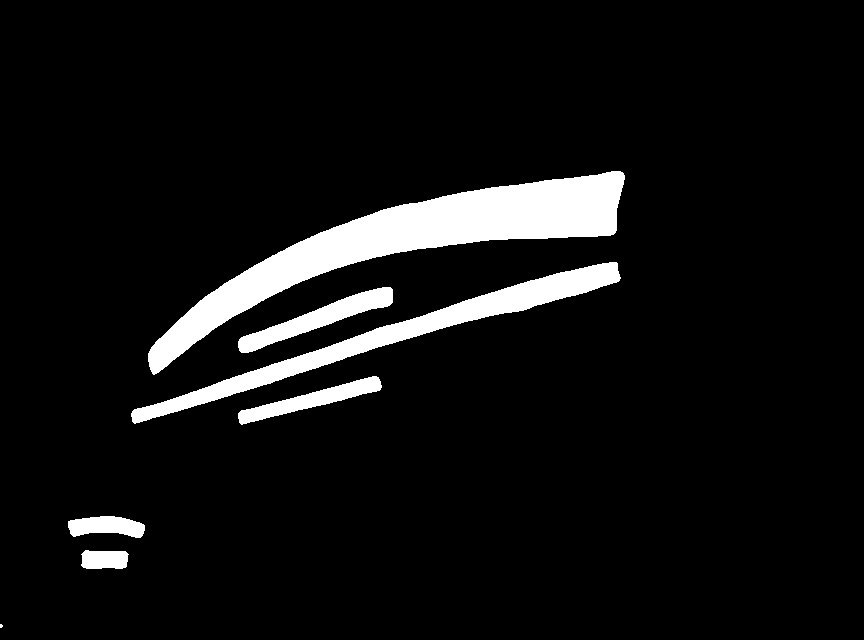}
		\caption{Text kernel map}
		\label{subfig:textkernel}
	\end{subfigure}
	\begin{subfigure}{.35\linewidth}
		\centering
		\includegraphics[width=\linewidth]{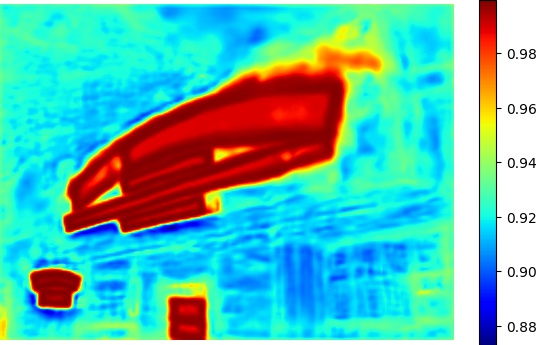}
		\caption{Relation matrix visualization}
		\label{subfig:global}
	\end{subfigure}
	\hspace{15pt}
	\begin{subfigure}{.35\linewidth}
		\centering
		\includegraphics[width=\linewidth]{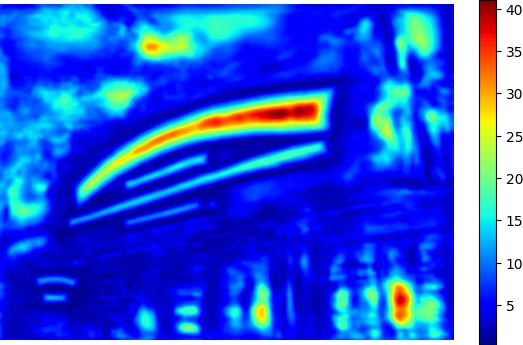}
		\caption{Distance map visualization}
		\label{subfig:local}
	\end{subfigure}
	\caption{Visualization of the detection result, enhanced text region map, text kernel map, relation matrix, and distance map.}
	\label{fig:matrixvisual}
\end{figure*}

\subsubsection{Multi-oriented Text Detection}
We also evaluate the proposed CBN on MSRA-TD500, ICDAR 2015, and COCO-Text test sets to verify the superiority in handling multi-oriented texts. During testing, we keep the aspect ratio of the test images and resize the shorter side of the images to 736 for MSRA-TD500, and the longer side of the images in ICDAR 2015 and COCO-Text are resized to 1280. As shown in Tab.~\ref{tab:msra} and Tab.~\ref{tab:cocotext}, our method achieves the F-measure of 89.7\% (MSRA-TD500) and 64.4\% (COCO-Text), surpassing all other state-of-the-art methods. This demonstrates the superiority of our method for detecting texts with various orientations and extreme aspect ratios in complex natural scenarios (see Fig.~\ref{fig:visualshowb}). For ICDAR 2015 dataset shown in Tab.~\ref{tab:icdar15}, although there is a slight gap (0.7\%) between our result and other SOTA methods, our model outperforms Raisi \emph{et al.}'s method~\cite{raisi2022arbitrary} and other methods under lightweight backbone.

\subsubsection{Multi-lingual Text Detection}
To demonstrate the robustness of our model for different languages, we test our model on two large-scale scene text datasets (\emph{i.e.} the MTWI test set and the ICDAR-MLT19 test set). During testing, we set both sides of the image in the range of (640, 1280) on MTWI and (960, 1920) on ICDAR-MLT19 respectively while keeping its aspect ratio. As shown in Tab.~\ref{tab:mtwi}, our method outperforms all of the counterparts by at least 1.6\% (76.8\% \emph{vs} 75.2\%) on MTWI test set. For ICDAR-MLT19 test set shown in Tab.~\ref{tab:mlt19}, our model achieves 5.2\% improvement in terms of F-measure when compared with the state-of-the-art model DBNet++. Notably, when using the lightweight backbone ResNet18 at the same time, the F-measure of our model is 5.4\% better than DBNet++. We also visualize our detection results in Fig.~\ref{fig:visualshowb}, which shows that our method can handle multi-lingual texts well.

\begin{table}[t!]
	\centering
	\caption{Quantitative detection results on MTWI test set. * and $\dagger$ indicate that the results are reported by SegLink++~\cite{tang2019seglink++} and MOST~\cite{he2021most} respectively. ``P", ``R", and ``F" represent precision, recall, and F-measure respectively.}
	\resizebox{\linewidth}{!}{
		\begin{tabular}{c|c|c|c|c}
			\hline
			{\textbf{Method}} &\textbf{Backbone} & \textbf{P(\%)} & \textbf{R(\%)}  & \textbf{F(\%)}   \\
			\hline
			\hline
			SegLink * ~\cite{shi2017detecting}   &VGG16 &70.0 &65.4 &67.6 \\
			TextBoxes++ * ~\cite{liao2018textboxes++} &VGG16  & 66.8 &56.3 &61.1  \\
			SegLink++~\cite{tang2019seglink++}  &VGG16  & 74.7 &69.7 &72.1  \\
			BDN $\dagger$ ~\cite{liu2019omnidirectional}   &Res50&77.3 &70.0 &73.4  \\
			PAN $\dagger$ ~\cite{wang2019efficient}    &Res50 &78.9 &68.9 &73.5  \\
			MOST~\cite{he2021most}    & Res50 &78.8 &71.1 &74.7  \\
			Tang \emph{et al.}~\cite{tang2022few}   &Res50  &78.4 &\textbf{72.3} &75.2  \\
			\hline
			\textbf{Ours}     & Res18 &\textbf{84.6} & 68.2 &75.5 \\
			\textbf{Ours}     &Res50 &82.1 &72.2 &\textbf{76.8} \\
			\textbf{Ours}     &Res50-DCN &83.8 &70.8 &76.7 \\
			\hline
		\end{tabular}
	}
	\label{tab:mtwi}
\end{table}

\begin{table}[t!]
	\centering
	\caption{Quantitative detection results on ICDAR-MLT19 test set. CRAFTS* used character-level annotations and integrated a recognition model. ``P", ``R", and ``F" represent precision, recall, and F-measure respectively.}
	\resizebox{\linewidth}{!}{
		\begin{tabular}{c|c|c|c|c}
			\hline
			{\textbf{Method}}  &\textbf{Backbone} & \textbf{P(\%)} & \textbf{R(\%)}  & \textbf{F(\%)}   \\
			\hline
			\hline
			PSENet~\cite{wang2019shape}  &Res50  &73.5 &59.6 &65.8  \\
			CRAFTS *~\cite{baek2019character}  &VGG16  &79.5 &59.6 &68.1 \\
			DBNet ~\cite{liao2020real}  &Res18-DCN &75.3 &60.2 &66.9  \\
			DBNet~\cite{liao2020real}  &Res50-DCN &78.3 &64.0 &70.4 \\
			DBNet++~\cite{liao2022real}  &Res18-DCN  &77.5 &61.0 &68.2 \\
			DBNet++~\cite{liao2022real}   &Res50-DCN &78.6 &65.4 &71.4 \\
			\hline
			\textbf{Ours}  &Res18   & 81.3&67.3 &73.6 \\
			\textbf{Ours}   &Res50  & 82.6 &69.5 &75.5 \\
			\textbf{Ours}   &Res50-DCN  & \textbf{84.8} &\textbf{69.7} &\textbf{76.6} \\
			\hline
		\end{tabular}
	}
	\label{tab:mlt19}
\end{table}

\subsubsection{Discussion about Model Complexity}
In this part, we discuss the model complexity of the proposed CBN, including parameters, computation, and inference speed. Firstly, as shown in Fig.~\ref{subfig:params} and Fig.~\ref{subfig:gflops}, Although the baseline model PAN~\cite{wang2019efficient} and PCR~\cite{dai2021progressive} have the lowest number of parameters and GFLOPs respectively, the F-measure of PAN is only 83.5\% on CTW1500, and the inference speed of PCR is limited by its NMS post-processing. Our proposed CBN brings considerable performance improvements of over 2\% F-measure on CTW1500 dataset with only about 0.01M parameters increase and 0.65\% GFLOPs increase when compared with the PAN baseline, which proves that CBN is an efficient lightweight plug-and-play for segmentation-based detectors. The comparison results of speed between our method and some previous methods can be seen in Tab.~\ref{tab:fpscomparesion} and Fig~\ref{subfig:speed}. When testing on CTW-1500, our method surpasses TextBPN~\cite{zhang2021adaptive} and DBNet~\cite{liao2020real} both on F-measure and speed (\emph{i.e.}, 86.0\% \emph{vs} 83.4\% and 85.0\% and 28.2 FPS \emph{vs} 6.8 FPS and 17.8 FPS). Compared with the baseline model PAN~\cite{wang2019efficient}, our method outperforms it by a large margin on F-measure (\emph{i.e.}, 85.7\% \emph{vs} 83.5\%), and only lags 0.1 FPS behind on speed (\emph{i.e.}, 30.7 FPS \emph{vs} 30.8 FPS). Overall, our method not only outperforms some previous methods in terms of F-measure but also keeps a real-time inference speed.

\begin{table}[t]
	\centering
	\caption{Comparision results of speed between our method and some previous methods on CTW1500 dataset.}
	\resizebox{\linewidth}{!}{
		\begin{tabular}{c|c|c|c}
			\hline
			{\textbf{~~Method~~}}    &\textbf{Backbone} &\textbf{F-measure} & \textbf{FPS} \\
			\hline
			\hline
			PSENet~\cite{wang2019shape}   &Res50&82.2  &6.1\\
			PAN~\cite{wang2019efficient}  &Res18 &83.5 &\textbf{30.8} \\
			TextBPN~\cite{zhang2021adaptive}  &Res50 &85.0 &6.8 \\
			PCR~\cite{dai2021progressive} &DLA34 &84.7 &9.5 \\
			TextBPN++~\cite{zhang2022arbitrary}   &Res18 &84.5 &26.2 \\
			TextBPN++~\cite{zhang2022arbitrary}   &Res50 &85.3 &9.4 \\
			DBNet~\cite{liao2020real}   &Res50-DCN &83.4 &17.8 \\
			DBNet++~\cite{liao2022real}    &Res50-DCN &85.3 &16.4 \\
			\hline
			\textbf{Ours-4s}     &Res18 &85.7 &30.7 \\
			\textbf{Ours-1s}      &Res18 &\textbf{86.0} &28.2 \\
			\hline
		\end{tabular}
	}
	\label{tab:fpscomparesion}
\end{table}

\begin{figure*}[t!]
	\centering
	\begin{subfigure}[b]{.3\textwidth}
		\centering
		\includegraphics[width=\textwidth]{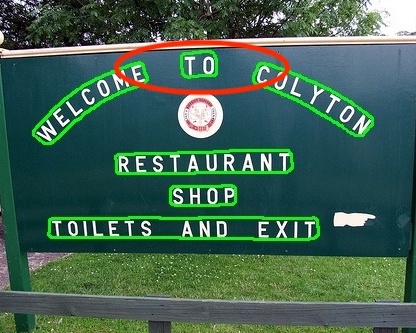}
		\caption{}
		\label{subfig:space}
	\end{subfigure}
	\hspace{-1.5pt}	
	\begin{subfigure}[b]{.3\textwidth}
		\centering
		\includegraphics[width=\textwidth]{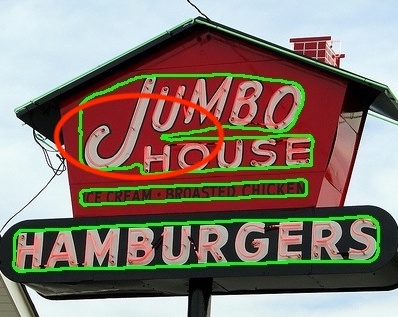}
		\caption{}
		\label{subfig:near}
	\end{subfigure}
	\hspace{-1.5pt}	
	\begin{subfigure}[b]{.3\textwidth}
		\centering
		\includegraphics[width=\textwidth]{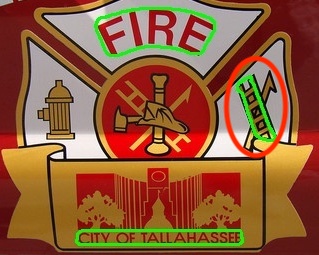}
		\caption{}
		\label{subfig:symbols}
	\end{subfigure}
	\caption{Failure samples. The green polygons are our detection results and the red line areas represent the failure part.}
	\label{fig:failure}
\end{figure*}

\subsubsection{Visualization of Context-aware Module}
In this part, we show the relation matrix (see Eq.~\ref{equ:grelation}) of the global text-contextual and the distance map (see the input $\textbf{D}$ in Eq.~\ref{equ:matlocal}) of the local text-contextual respectively. The relation matrix describes the global relationship between each pixel and all text representations. Fig.~\ref{subfig:textregion} and \ref{subfig:textkernel} show the enhanced text region and text kernel map respectively. We can find that the relation matrix of the global text-contextual has high response values in the text area, and these response values on each text line are close or similar (see Fig.~\ref{subfig:global}). The distance map represents the relationship between each text pixel and its corresponding text instance. Since different text lines have different distance values between the text pixel and its text boundary, the response values in the distance map are gradually changing in the text area (see Fig.~\ref{subfig:local}). There are two roles in the distance map. On the one hand, the distance map describes the nearest distance between each positive pixel and its corresponding text boundary, which is used to rebuild the complete text instances adaptively. On the other hand, the normalized distance map provides a relation matrix for the local text-contextual module. Thanks to the relation matrix and the distance map for the global and local context-aware modules, CBN can establish the global and local relationship between pixels and learn more comprehensive text-contextual.

\subsection{Limitation}
As demonstrated in previous experiments, the proposed CBN works well in most cases of arbitrary-shaped text detection. It still fails for some difficult cases as shown in Fig.~\ref{fig:failure}. On the one hand, whether two adjacent text instances should be considered as one text instance or not is still a challenging problem, which influences the detection performance deeply (see Fig.~\ref{subfig:space} and \ref{subfig:near}). On the other hand, our method may mistakenly treat some decorative patterns as texts and produce false positives (see Fig.~\ref{subfig:symbols}). For these two problems, it is important and useful to add some high-level semantic information for joint training of the model. In the future, we will plan to solve these problems and make the model more robust.

\section{Conclusion}
In this paper, we propose a novel network named CBN, containing a context-aware module and a boundary-guided module, to remedy the drawbacks of segmentation-based text detection methods. Different from previous methods, which predict each pixel independently, the proposed context-aware module encodes the relationship between pixels into global and local contexts to enhance the text kernel segmentation results. To expand the text kernel quickly and accurately, we propose a boundary-guided module to guide the text kernel contours towards the text boundaries by using a learnable distance map, which leads to a favorable accuracy-speed trade-off. Moreover, the proposed lightweight CBN can be readily plugged into several segmentation-based text detectors. Our method achieves state-of-the-art performance on several datasets, including curve texts, multi-oriented texts, and multi-lingual texts. Ablation studies show that our proposed context-aware module can improve the performance of segmentation-based text detectors, and the boundary-guided module can achieve a good balance between speed and performance.

\bibliography{sn-bibliography}


\end{document}